# The Past, Current, and Future of Neonatal Intensive Care Units with Artificial Intelligence: A Systematic Review

**Elif Keles, MD, PhD[1] *, Ulas Bagci, PhD[1,2,3]**

Elif Keles, MD, PhD[1]

Northwestern University, Feinberg School of Medicine, Department of Radiology, Chicago, IL[1]

Ulas Bagci, PhD[1,2,3]

Northwestern University, Feinberg School of Medicine, Department of Radiology, Chicago, IL [1], Northwestern University, Department of Biomedical Engineering [2], and [3] Department of Electrical and Computer Engineering [3]

*: Corresponding Author

elif.keles@northwestern.edu

**Abstract**

Machine learning and deep learning are two subsets of artificial intelligence that involve teaching computers to learn and make decisions from any sort of data. Most recent developments in artificial intelligence are coming from deep learning, which has proven revolutionary in almost all fields, from computer vision to health sciences. The effects of deep learning in medicine have changed the conventional ways of clinical application significantly. Although some sub-fields of medicine, such as pediatrics, have been relatively slow in receiving the critical benefits of deep learning, related research in pediatrics has started to accumulate to a significant level, too.

Hence, in this paper, we review recently developed machine learning and deep learning-based solutions for neonatology applications. We systematically evaluate the roles of both classical machine learning and deep learning in neonatology applications, define the methodologies, including algorithmic developments, and describe the remaining challenges in the assessment of neonatal diseases by using PRISMA 2020 guidelines. To date, the primary areas of focus in neonatology regarding AI applications have included survival analysis, neuroimaging, analysis of vital parameters and biosignals, and retinopathy of prematurity diagnosis. We have categorically summarized **106** research articles from 1996 to 2022 and discussed their pros and cons, respectively. In this systematic review, we aimed to further enhance the comprehensiveness of the study. We also discuss possible directions for new AI models and the future of neonatology with the rising power of AI, suggesting roadmaps for the integration of AI into neonatal intensive care units.

## Introduction

The AI tsunami fueled by advances in artificial intelligence (AI) is constantly changing almost all fields, including healthcare; it is challenging to track the changes originated by AI as there is not a single day that AI is not applied to anything new. While AI affects daily life enormously, many clinicians may not be aware of how much of the work done with AI technologies may be put into effect in today's healthcare system.

In this review, we fill this gap, particularly for physicians in a relatively underexplored area of AI: neonatology. The origins of AI, specifically machine learning (ML), can be tracked all the way back to the 1950s, when Alan Turing invented the so-called "learning machine" as well as military applications of basic AI[1]. During his time, computers were huge, and the cost of increased storage space was astronomical. As a result, their capabilities, although substantial for their day, were restricted. Over the decades, incremental advancements in theory and technological advances steadily increased the power and versatility of ML[2].

How do machine learning (ML) and deep learning(DL) work? ML falls under the category of AI[2]. ML's capacity to deal with data brought it to the attention of computer scientists. ML algorithms and models can learn from data, analyze, evaluate, and make predictions or decisions based on learning and data characteristics. DL is a subset of ML. Different from this larger class of ML definitions, the underlying concept of DL is inspired by the functioning of the human brain, particularly the neural networks responsible for processing and interpreting information. DL mimics this operation by utilizing artificial neurons in a computer neural network. In simple terms, DL finds weights for each artificial neuron that connects to each other from one layer to another layer. Once the number of layers is high (i.e., deep), more complex relationships between input and output can be modeled[3-5]. This enables the network to acquire more intricate representations of the data as it learns. The utilization of a hierarchical approach enables DL models to autonomously extract features from the data, as opposed to depending on human-engineered features as is customary in conventional ML[3]. DL is a highly specialized form of ML that is ideally modified for tasks involving unstructured data, where the features in the data may be learnable, and exploration of non-linear associations in the data can be possible[6-8].

The main difference between ML and DL lies in the complexity of the models and the size of the datasets they can handle. ML algorithms can be effective for a wide range of tasks and can be relatively simple to train and deploy[6,7,9-11]. DL algorithms, on the other hand, require much larger datasets and more complex models but can achieve exceptional performance on tasks that involve high-dimensional, complex data[7]. DL can automatically identify which aspects are significant, unlike classical ML, which requires pre-defined elements of interest to analyze the data and infer a decision[10]. Each neuron in DL architectures (i.e., artificial neural networks(ANN)) has non-linear activation function(s) that help it learn complex features representative of the provided data samples[9].

ML algorithms, hence, DL, can be categorized as either supervised, unsupervised, or reinforcement learning based on the input-output relationship. For example, if output labels (outcome) are fully available, the algorithm is called "supervised," while unsupervised algorithms explore the data without their reference standards/outcomes/labels in the output [3,12]. In terms of applications, both DL and ML are typically used for tasks such as classification, regression, and clustering[6,9,10,13-15]. DL methods' success depends on the availability of large-scale data, new optimization algorithms, and the availability of GPUs[6,10]. These algorithms are designed to autonomously learn and develop as they gain experience, like humans[3]. As a result of DL's powerful representation of the data, it is considered today's most improved ML method, providing drastic changes in all fields of medicine and technology, and it is the driving force behind virtually all progress in AI today[5] **(Figure 1).**

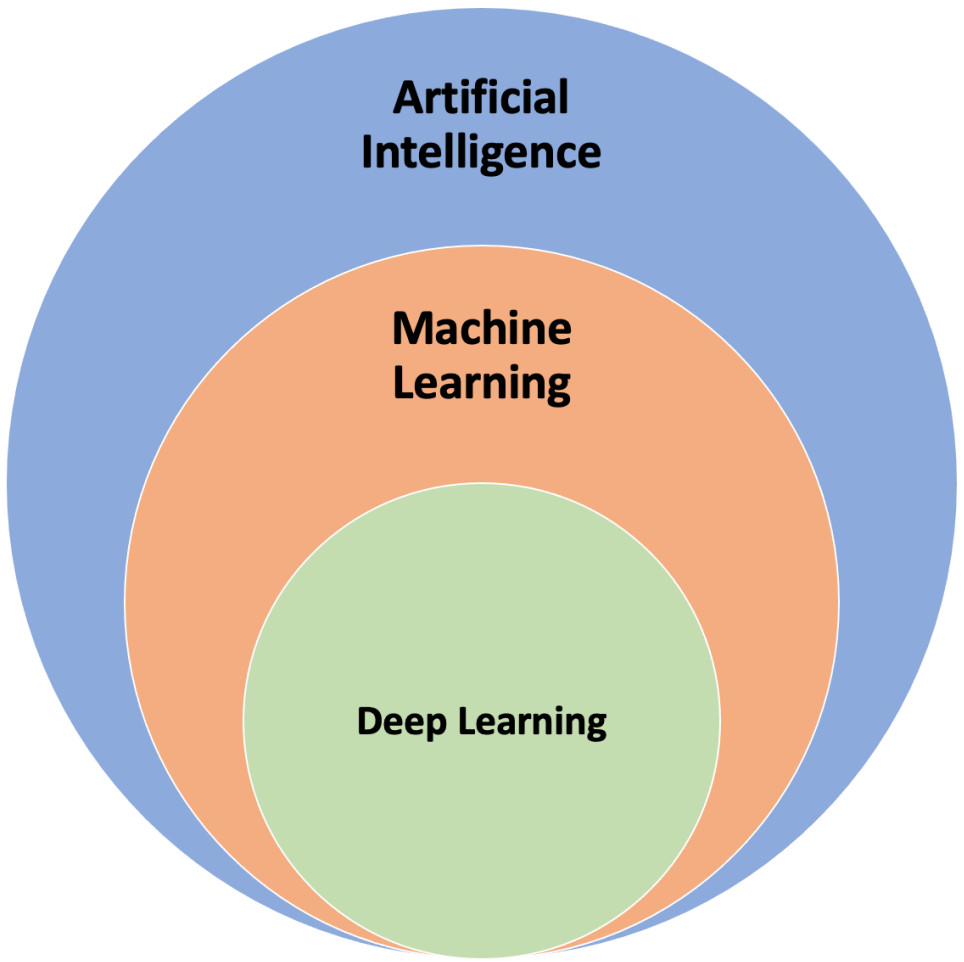

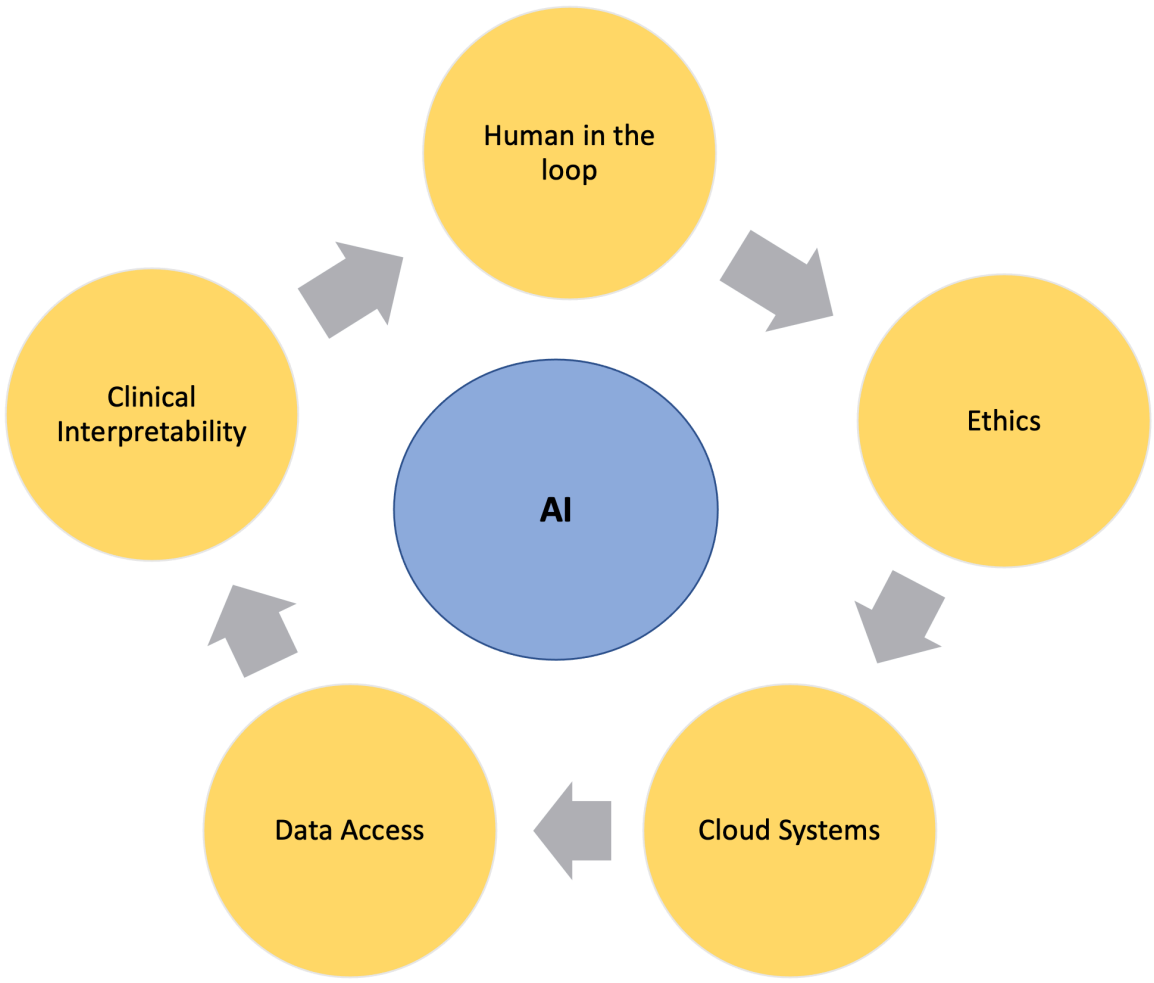

1

1b

**Figure 1a: Hierarchical diagram of AI**.
How do machine learning (ML) and deep learning(DL) work? ML falls under the category of AI. DL is a subset of ML.

**Figure 1b: Ongoing hurdles of AI when applied to healthcare applications.**
Key concerns related to AI and each concern affects the outcome of AI in Neonatology including;1) challenges with clinical interpretability; 2) knowledge gaps in decision-making mechanisms, with the latter requiring human-in-the-loop systems
3) ethical considerations;
4) the lack of data and annotations, and 5) the absence of Cloud systems allowing for secure data sharing and data privacy.

There are three major problem types in DL in medical imaging: image segmentation, object detection (i.e., an object can be an organ or any other anatomical or pathological entity), and image classification (e.g., diagnosis, prognosis, therapy response assessment) [3]. Several DL algorithms are frequently employed in medical research; briefly, those approaches belong to the following family of algorithms:

Convolutional Neural Networks (CNNs) are predominantly employed for tasks related to computer vision and signal processing. CNNs can handle tasks requiring spatial relationships where the columns and rows are fixed, such as imaging data. CNN architecture encompasses a sequence of phases (layers) that facilitate the acquisition of hierarchical features. Initial phases (layers) extract more local features such as corners, edges, and lines, later phases (layers) extract more global features. Features are propagated from one layer to another layer, and feature representation becomes richer this way. During feature propagation from one layer to another layer, the features are added certain nonlinearities and regularizations to make the functional modeling of input-output more generalizable. Once features become extremely large, there are operations within the network architecture to reduce the feature size without losing much information, called *pooling* operations. The auto-generated and propagated features are then utilized at the end of the network architecture for prediction purposes (segmentation, detection, or classification) [3,16].

Recurrent Neural Networks (RNNs) are designed to facilitate the retention of sequential data, namely text, speech, and time-series data such as clinical data or electronic health records (EHRs). They can capture temporal relationships between data components, which can be helpful for predicting disease progression or treatment outcomes[11,17,18]. RNNs use similar architecture components that CNNs have. Long Short-Term Memory (LSTM) models are types of RNNs and are commonly used to overcome their shortcomings because they can learn long-term dependencies in data better than conventional RNN architectures. They are utilized in some classification tasks, including audio[17,19]. LSTM utilizes *a gated memory cell* in the network architecture to store information from the past; hence, the memory cell can store information for a long period of time, even if the information is not immediately relevant to the current task. This allows LSTMs to learn patterns in data that would be difficult for other types of neural networks to learn.

Generative adversarial networks (GANs) are a class of DL models that can be used to generate new data that is like existing data. In healthcare, GANs have been used to generate synthetic medical images. There are two CNNs (generator and discriminator); the first CNN is called the generator, and its primary goal is to make synthetic images that mimic actual images. The second CNN is called the discriminator, and its main objective is to identify between artificially generated images and real images[20]. The generator and discriminator are trained jointly in a process called adversarial training, where the generator tries to create data that is so realistic that the discriminator cannot distinguish it from real data. GANs are used to generate a variety of different types of data, including images, videos, and text. GANs are used to enhance image quality, signal reconstruction, and other tasks such as classification and segmentation too[20-22].

Transfer learning (TL) is a concept derived from cognitive science that states that information is transferred across related activities to improve performance on a new task. It is generally known that people can accomplish similar tasks by building on prior knowledge[23]. TL has been implemented to minimize the need for annotation by transferring DL models with knowledge from a previous task and then fine-tuning them in the current task[24]. The majority of medical image classification techniques employ TL from pretrained models, such as *ImageNet*, which has been demonstrated to be inefficient due to the ImageNet consisting of natural images[25]. The approaches that utilized *ImageNet* pre-trained images in CNNs revealed that fine-tuning more layers provided increased accuracy[26]. The initial layers of ImageNet-pretrained networks, which detect low-level image characteristics, including corners and borders, may not be efficient for medical images[25,26].

New and more advanced DL algorithms are developed almost daily. Such methods could be employed for the analysis of imaging and non-imaging data in order to enhance performance and reliability. These methods include Capsule Networks, Attention Mechanisms, and Graph Neural Networks (GNNs) [27-30]. Briefly, these are:

Capsule Networks are a relatively new form of DL architecture that aim to address some of the shortcomings of CNNs: pooling operations (reducing the data size) and a lack of hierarchical relations between objects and their parts in the data. Capsules can capture spatial relationships between features and are more capable of handling rotations and deformations of image objects thanks to their vectorial representations in neuronal space. Capsule Networks have shown potential in image classification tasks and could have applications in medical imaging analysis[27]. However, its implementation and computational time are two hurdles that restrict its widespread use.

Attention Mechanisms, represented by Transformers, have contributed to the development of computer vision and language processing. Unlike CNNs or RNNs, transformers allow direct interaction between every pair of components within a sequence, making them particularly effective at capturing long-term relationships[29,30]. More specifically, a self-attention mechanism in Transformers is an important piece of the DL model as it can dynamically focus on different parts of the input data sequence when producing an output, providing better context understanding than CNN based systems.

Graph Neural Networks (GNNs) are a form of data structure that describes a collection of objects (nodes) and their relationships (edges). There are three forms of tasks, including node-level, edge-level, and graph level[31]. Graphs may be used to denote a wide range of systems, including molecular interaction networks, and bioinformatics[31-33]. GNNs have demonstrated potential in both imaging and non-imaging data analysis[28,34].

Physics-driven systems are needed in imaging field. Several studies have demonstrated the effectiveness of DL methods in the medical imaging field[35-39]. As the field of DL continues to evolve, it is likely that new methods and architectures will emerge to address the unique challenges and constraints of various types of data. One of the most common problems faced with DL based MRI construction[35]. Specific algorithms for this problem can be essentially categorized into two groups: data driven and physics driven algorithms. In purely data-driven approaches, a mapping is learned between the aliased image and the image without artifacts[39]. Acquiring fully sampled (artifact-free) data sets is impractical in many clinical imaging studies when organs are in motion, such as the heart, and lung. Recently developed models can employ these under sampled MRI acquisitions as input and generate output images consistent with fully-sampled (artifact free) acquisitions[37-39].

What is the Hybrid Intelligence? A highly desirable way of incorporating advances in AI is to let AI and human intellect work together to solve issues, and this is referred to as "hybrid intelligence"[40] (e.g., one may call this “mixed intelligence” or “human-in-the-loop AI systems”). This phenomenon involves the development of AI systems that serve to supplement and amplify human decision-making processes, as opposed to completely replacing them[3]. The concept involves integrating the respective competencies of artificial intelligence and human beings in order to attain superior outcomes that would otherwise be unachievable [41]. AI algorithms possess the ability to process extensive amounts of data, recognize patterns, and generate predictions rapidly and precisely. Meanwhile, humans can contribute their expertise, understanding, and intuition to the discussion to offer context, analyze outcomes, and render decisions[42]. The hybrid intelligence strategy can help decision-makers in a variety of fields make decisions that are more precise, effective, and efficient by combining these qualities[3,4,43,44]. Human in the loop and hybrid intelligence systems are promising for time consuming tasks in healthcare and neonatology.

Where do we stand currently? AI in medicine has been employed for over a decade, and it has often been considered that clinical implementation is not completely adapted to daily practice in most of the clinical field[5,45,46]. In recent years, increasingly complex computer algorithms and updated hardware technologies for processing and storing enormous data sets have contributed to this achievement[6,7,46,47]. It has only been within the last decade that these systems have begun to display their full potential[6,9]. The field of AI research appears to have been taken up with differing degrees of enthusiasm across disciplines. When analyzing the thirty years of research into AI, DL, and ML conducted by several medical subfields between the years 1988 and 2018, one-third of publications in DL yielded to radiology, and most of them are within the imaging sciences (radiology, pathology, and cell imaging) [48]. Software systems work by utilizing biomedical images with predictive/diagnostic/prognostic features and integrating clinical or pre-clinical data. These systems are designed with ML algorithms[46]. Such breakthrough methods in DL are nowadays extensively applied in pathology, dermatology, ophthalmology, neurology, and psychiatry[6,47,49]. AI has its own difficulties with the increasing utilization of healthcare (**Figure 1b**).

What are the needs in clinics? Clinicians are concerned about the healthcare system's integration with AI: there is an exponential need for diagnostic testing, early detection, and alarm tools to provide diagnosis and novel treatments without invasive tests and procedures[50]. Clinicians have higher expectations of AI in their daily practices than before. AI is expected to decrease the need for multiple diagnostic invasive tests and increase diagnostic accuracy with less invasive (or non-invasive) tests. Such AI systems can easily recognize imaging patterns on test images (i.e., unseen or not utilized efficiently in daily routines), allowing them to detect and diagnose various diseases. These methods could improve detection and diagnosis in different fields of medicine.

The overall goal of this systematic review is to explain AI's potential use and benefits in the field of neonatology. We intend to enlighten the potential role of AI in the future in neonatal care. We postulate that AI would be best used as a hybrid intelligence (i.e., human-in-the-loop or mixed intelligence) to make neonatal care more feasible, increase the accuracy of diagnosis, and predict the outcome and diseases in advance. The rest of the paper is organized as follows: In results, we explain the published AI applications in neonatology along with AI evaluation metrics to fully understand their efficacy in neonatology and provide a comprehensive overview of DL applications in neonatology. In discussion, we examine the difficulties of AI utilization in neonatology and future research discussions. In the methods section, we outline the systematic review procedures, including the examination of existing literature and the development of our search strategy.

We review the past, current, and future of AI-based diagnostic and monitoring tools that might aid neonatologists' patient management and follow-up. We discuss several AI designs for electronic health records, image, and signal processing, analyze the merits and limits of newly created decision support systems, and illuminate future views clinicians and neonatologists might use in their normal diagnostic activities. AI has made significant breakthroughs to solve issues with conventional imaging approaches by identifying clinical variables and imaging aspects not easily visible to human eyes. Improved diagnostic skills could prevent missed diagnoses and aid in diagnostic decision-making. The overview of our study is structured as illustrated in **Figure 2**. Briefly, our objectives in this systematic review are:

- to explain the various AI models and evaluation metrics thoroughly explained and describe the principal features of the AI models,
- to categorize neonatology-related AI applications into macro-domains, to explain their sub-domains and the important elements of the applicable AI models,
- to examine the state-of-the-art in studies, particularly from the past several years, with an emphasis on the use of ML in encompassing all neonatology,
- to present a comprehensive overview and classification of DL applications utilized and in neonatology,
- to analyze and debate the current and open difficulties associated with AI in neonatology, as well as future research directions, to offer the clinician a comprehensive perspective of the actual situation.

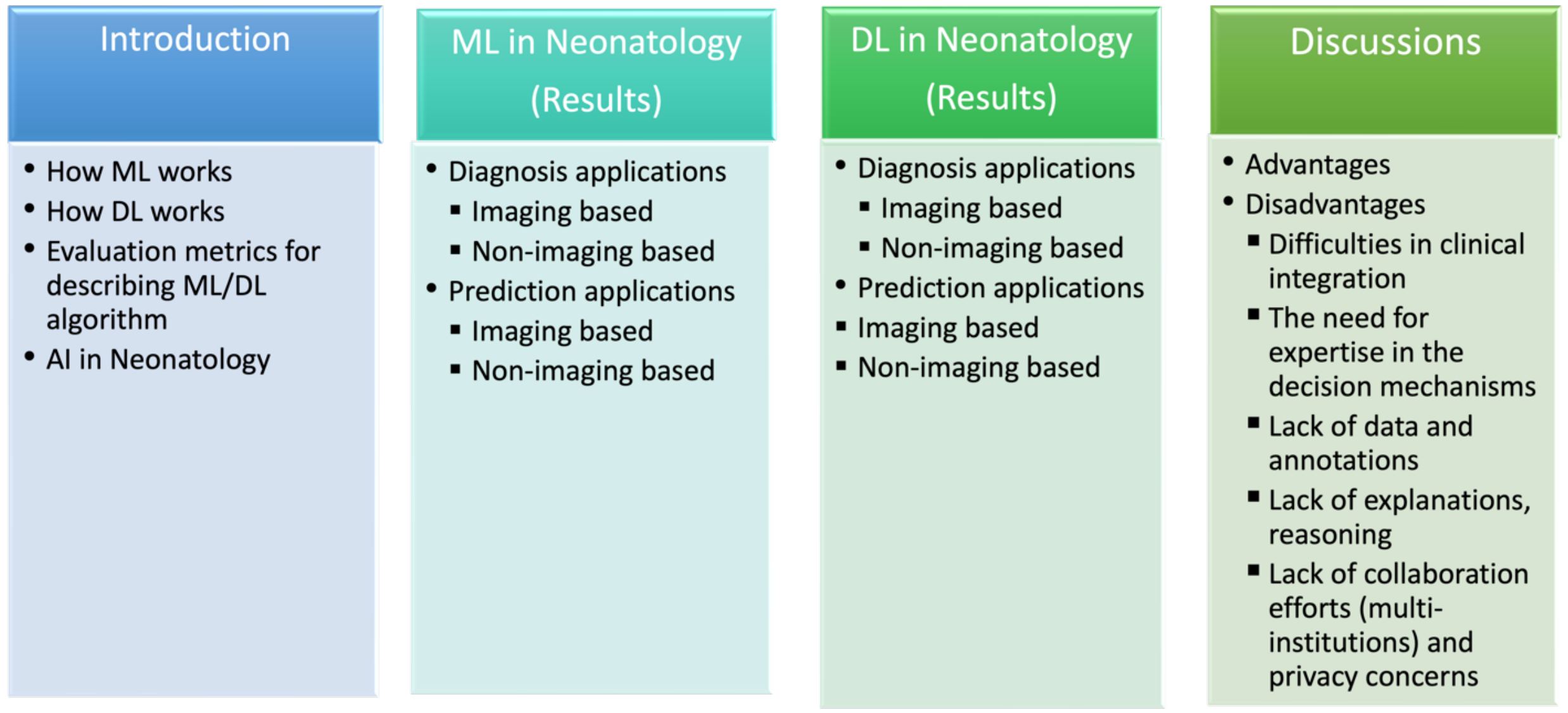

**Figure 2:** An overview of the structure of this paper.
It is provided an overview of our paper's structure and objectives:

1. Explaining AI Models and Evaluation Metrics:
2. Evaluating ML applied studies in Neonatology
3. Evaluating DL applied studies in Neonatology
4. Analyzing Challenges and Future Directions

AI covers a broad concept for the application of computing algorithms that can categorize, predict, or generate valuable conclusions from enormous data sets[46]. Algorithms such as Naive Bayes, Genetic Algorithms, Fuzzy Logic, Clustering, Neural Networks(NN), Support Vector Machines(SVM), Decision Trees, and Random Forests(RF) have been used for more than three decades for detection, diagnosis, classification, and risk assessment in medicine as ML methods[9,10]. Conventional ML approaches for image classification involve using hand-engineered features, which are visual descriptions and annotations learned from radiologists, that are encoded into algorithms.

Images, signals, genetic expressions, EHR, and vital signs are examples of the various unstructured data sources that comprise medical data **(Figure 3)**. Due to the complexity of their structures, DL frameworks may take advantage of this heterogeneity by attaining high abstraction levels in data analysis.

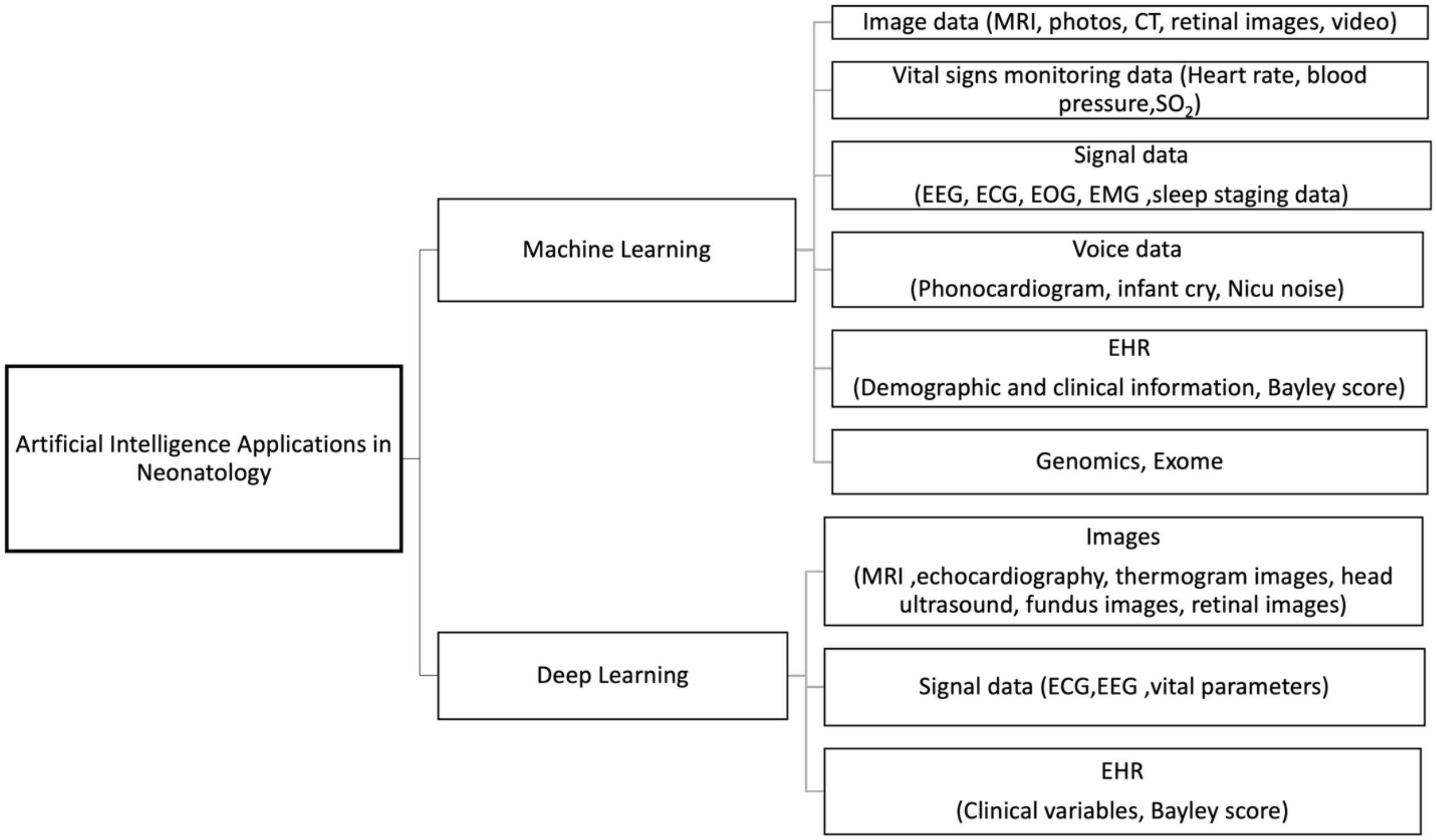

**Figure 3:** An overview of AI applications in neonatology.
Unstructured data such as medical images, vital signals, genetic expressions, EHRs, and signal data contribute to the wide variety of medical information. Analyzing and interpreting different data streams in neonatology requires a comprehensive strategy because each has unique characteristics and complications.

While ML requires manual/hand-crafted selection of information from incoming data and related transformation procedures, DL performs these tasks more efficiently and with higher efficacy[9,10,46]. DL is able to discover these components by analyzing a large number of samples with a high degree of automation[7]. The literature on these ML approaches is extensive before the development of DL[5,7,45].

It is essential for clinicians to understand how the suggested ML model should enhance patient care. Since it is impossible for a single metric to capture all the desirable attributes of a model, it is customarily necessary to describe the performance of a model using several different metrics. Unfortunately, many end-users do not have an easy time comprehending these measurements. In addition, it might be difficult to objectively compare models from different research models, and there is currently no method or tool available that can compare models based on the same performance measures[51]. In this part, the common ML and DL evaluation metrics are explained so neonatologists could adapt them into their research and understand of upcoming articles and research design [51,52].

AI is commonly utilized everywhere, from daily life to high-risk applications in medicine. Although slower compared to other fields, numerous studies began to appear in the literature investigating the use of AI in neonatology. These studies have used various imaging modalities, electronic health records, and ML algorithms, some of which have barely gone through the clinical workflow. Though there is no systematic review and future discussions in particular in this field[53-55]. Many studies were dedicated to introducing these systems into neonatology. However, the success of these studies has been limited. Lately, research in this field has been moving in a more favorable direction due to exciting new advances in DL. Metrics for evaluations in those studies were the standard metrics such as sensitivity (true-positive rate), specificity (true-negative rate), false-positive rate, false-negative rate, receiver operating characteristics (ROC), area under the ROC curves

(AUC), and accuracy **(Table 1).**

**Table 1: Evaluation metrics in artificial intelligence.**

| Term | Definition |
|---|---|
| True Positive (TP) | The number of positive samples that have been correctly identified. |
| True Negative (TN) | The number of samples that were accurately identified as negative. |
| False Positive (FP) | The number of samples that were incorrectly identified as positive. |
| False Negative (FN) | The number of samples that were incorrectly identified as negative. |
| Accuracy (ACC) | The proportion of correctly identified samples to the total sample count in the assessment dataset.<br>The accuracy is limited to the range [0, 1], where 1 represents properly predicting all positive and negative samples and 0 represents successfully predicting none of the positive or negative samples. |
| Recall (REC) | The sensitivity or True Positive Rate (TPR) is the proportion of correctly categorized positive samples to all samples allocated to the positive class. It is computed as the ratio of correctly classified positive samples to all samples assigned to the positive class. |
| Specificity (SPEC) | The negative class form of recall (sensitivity) and reflects the proportion of properly categorized negative samples. |
| Precision (PREC) | The ratio of correctly classified samples to all samples assigned to the class. |
| Positive Predictive Value (PPV) | The proportion of correctly classified positive samples to all positive samples. |
| Negative Predictive Value (NPV) | The ratio of samples accurately identified as negative to all samples classified as negative. |
| F1 score (F1) | The harmonic mean of precision and recall, which eliminates excessive levels of either. |
| Cross Validation | A validation technique often employed during the training phase of modeling, without no duplication among validation components. |
| AUROC (Area under ROC curve - AUC) | A function of the effect of various sensitivities (true-positive rate) on false-positive rate. It is limited to the range [0, 1], where 1 represents properly predicting all cases of all and 0 represents predicting the none of cases. |
| ROC | By displaying the effect of variable levels of sensitivity on specificity, it is possible to create a curve that illustrates the performance of a particular predictive algorithm, allowing readers to easily capture the algorithm's value. |
| Overfitting | Modeling failure indicating extensive training and poor performance on tests. |
| Underfitting | Modeling failure indicating inadequate training and inadequate test performance. |
| Dice Similarity Coefficient | Used for image analysis. It is limited to the range [0, 1], where 1 represents properly segmenting of all images and 0 represents successfully segmenting none of images. |

## Results

This systematic review was guided by the Preferred Reporting Items for Systematic Reviews and Meta-Analyses (PRISMA) protocol[56]. The search was completed on 11$^{st}$ of July 2022. The initial search yielded many articles (approximately 9000), and we utilized a systematic approach to identify and select relevant articles based on their alignment with the research focus, study design, and relevance to the topic. We checked the article abstracts, and we identified 987 studies. Our search yielded **106** research articles between **1996 and 2022 (Figure 4)**. Risk of bias summary analysis was done by the QUADAS-2 tool **(Figures 5 and 6)** [57-59]**.**

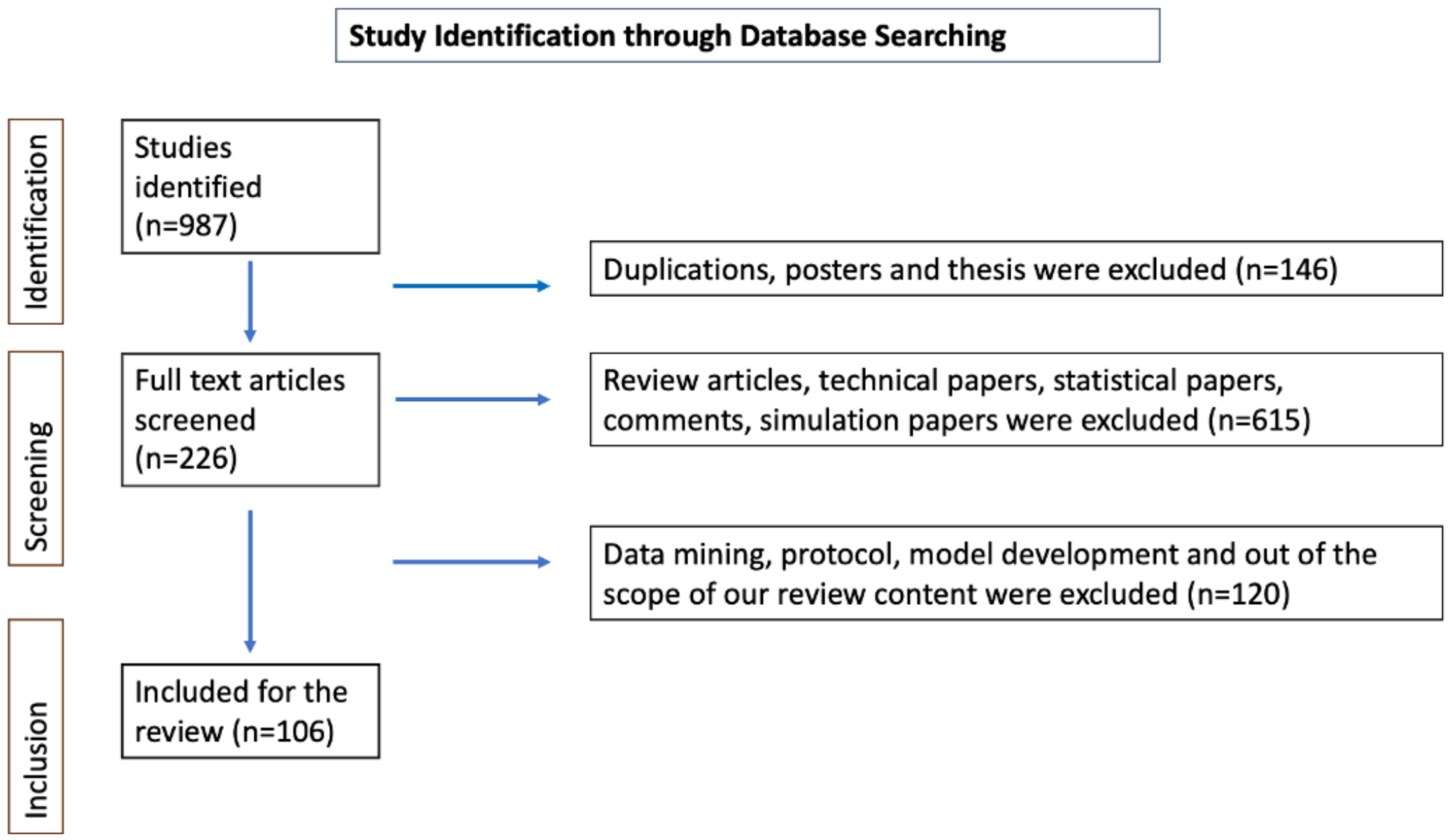

**Figure 4:** Identification of studies through database searches.

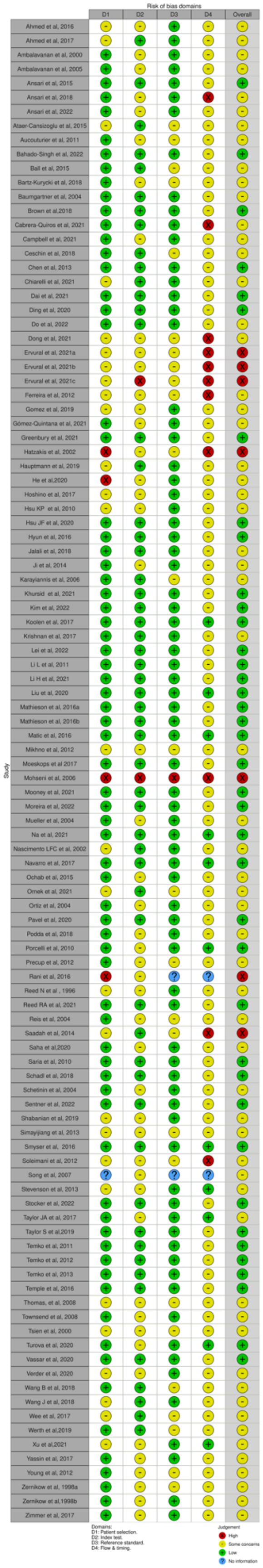

**Figure 5**: Bias summary of all research according to the QUADAS-2.
Risk of bias summary analysis was done by the QUADAS-2 tool.

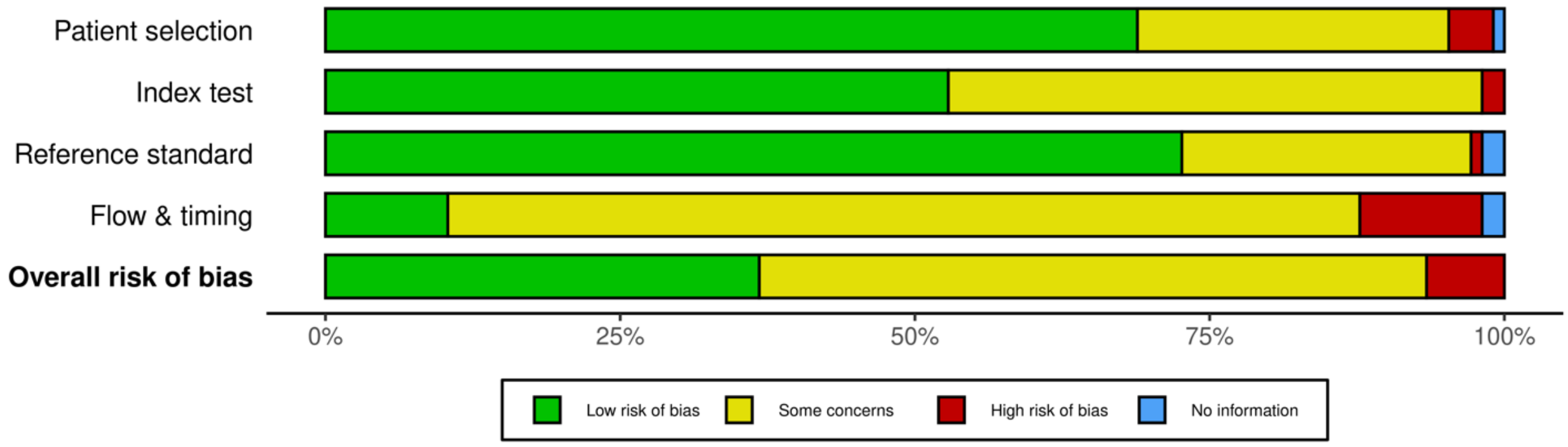

**Figure 6:** Bias summary of all studies according to the QUADAS-2.
Risk of bias summary analysis was done by the QUADAS-2 tool.

Our findings are summarized in two groups of tables: **Tables 2 - 5** summarize the AI methods from the pre-deep learning era **("Pre-DL Era")** in neonatal intensive care units according to the type of data and applications. **Tables 6 and 7**, on the other hand, include studies from the **DL Era**. Applications include classification (i.e., prediction and diagnosis), detection (i.e., localization), and segmentation (i.e., pixel level classification in medical images).

**Table 1:** ML based (non-DL) studies in neonatology using imaging data for diagnosis.

| Study | Approach | Purpose | Dataset | Type of data | Performance | Pros(+)<br>Cons(-) |
|---|---|---|---|---|---|---|
| Hoshino et al, 2017 [60] | CLAFIC, logistic regression analysis | To determine optimal color parameters predicting Biliary atresia (BA)stools | 50 neonates | 30 BA and 34 non-BA images | 100% (AUC) | +Effective and convenient modality for early detection of BA, and potentially for other related diseases<br>-Small sample size |
| Dong et al, 2021 [61] | Level Set algorithm | To evaluate the postoperative | 60 neonates | CT images | 84.7% (accuracy) | + Segmentation algorithm can accurately segment the CT image, so |
| | | enteral nutrition of neonatal high intestinal obstruction and analyze the clinical treatment effect of high intestinal obstruction | | | | that the disease location and its contour can be displayed more clearly.<br>- EHR(not included AI analysis)<br>-Small sample size<br>-Retrospective design |
| Ball et al, 2015 [62] | Random Forest (RF) | To compare whole-brain functional connectivity in preterm newborns at term-equivalent age with healthy term-born neonates in order to determine if preterm birth leads in particular changes to functional connectivity by term-equivalent age. | 105 preterm infants and 26 term controls | Both resting state functional MRI and T2-weighted Brain MRI | 80% (accuracy) | +Prospective<br>+Connectivity differences between term and preterm brain<br>-Not well-established model |
| Smyser et al, 2016 [63] | Support vector machine (SVM)-multivariate pattern analysis (MVPA) | To compare resting state-activity of preterm-born infants (Scanned at term equivalent postmenstrual age) to term infants | 50 preterm infants (born at 23–29 weeks of gestation and without moderate–severe brain injury) 50 term-born control infants studied | Functional MRI data<br>+<br>Clinical variables | 84% (accuracy) | +Prospective<br>+ GA at birth was used as an indicator of the degree of disruption of brain development<br>+ Optimal methods for rs-fMRI data acquisition and preprocessing for this population have not yet been rigorously defined<br>-Small sample size |
| Zimmer et al, 2017 [64] | NAF: Neighborhood approximation forest classifier of forests | To reduce the complexity of heterogeneous data population, manifold learning techniques are applied, which find a low-dimensional representation of the data. | 111 infants (NC, 70 subjects), affected by IUGR (27 subjects) or VM (14 subjects). | 3 T brain MRI | 80% (accuracy) | +Combining multiple distances related to the condition improves the overall characterization and classification of the three clinical groups (Normal, IUGR, Ventriculomegaly)<br>-The lack of neonatal data due to challenges during acquisition and data accessibility<br>-Small sample size |
| Krishnan et al, 2017 [65] | Unsupervised machine learning: | Variability in the Peroxisome Proliferator | 272 infants born at less than 33 wk gestational | Diffusion MR Imaging Diffusion Tractography | 63% (AUC) | + Inhibited brain development found in individuals exposed to the |
| | Sparse Reduced Rank Regression (sRRR) | Activated Receptor (PPAR) pathway would be related to brain development | age (GA) | Genome wide Genotyping | | stress of a preterm extrauterine world is controlled by genetic variables, and PPARG signaling plays a previously unknown cerebral function<br>-Further work is required to characterize the exact relationship between PPARG and preterm brain development, notably to determine whether the effect is brain specific or systemic |
| Chiarelli et al, 2021[66] | Multivariate statistical analysis | To better understand the effect of prematurity on brain structure and function, | 88 newborns | 3 Tesla BOLD and anatomical brain MRI<br><br>Few clinical variables | The multivariate analysis using motion information could not significantly infer GA at birth | +Prematurity was associated with bidirectional alterations of functional connectivity and regional volume<br>-Retrospective design<br>-Small sample size |
| Song et al, 2007 [67] | Fuzzy nonlinear support vector machines (SVM). | Neonatal brain tissue segmentation in clinical magnetic resonance (MR) images | 10 term neonates | Brain MRI T1 and T2 weighted | 70%-80%(dice score-gray matter)<br><br>65%-80% (dice score-white matter) | + Nonparametric modeling adapts to the spatial variability in the intensity statistics that arises from variations in brain structure and image inhomogeneity<br>+ Produces reasonable segmentations even in the absence of atlas prior<br>-Small sample size |
| Taylor et al, 2017 [68] | Machine Learning | Technology that uses a smartphone application has the potential to be a useful methodology for effectively screening newborns for jaundice | 530 newborns | Paired BiliCam images<br><br>total serum bilirubin (TSB) levels | High-risk zone TSB level was 95% for BiliCam and 92% for TcB ($P = .30$); for identifying newborns with a TSB level of ≥17.0, AUCs were 99% and 95%, respectively ($P = 0.09$). | + Inexpensive technology that uses commodity smartphones could be used to effectively screen newborns for jaundice<br>+Multicenter data<br>+Prospective design<br>-Method and algorithm name were not explained |
| Ataer-Cansizoglu et al, 2015 [69] | Gaussian Mixture Models | To develop novel | | 77 wide-angle retinal images | 95% | +Arterial and venous tortuosity (combined), and a large circular cropped image (with radius 6 times |
| | i-ROP | computer based image analysis system for grading plus diseases in ROP | | | (accuracy) | the disc diameter), provided the highest diagnostic accuracy<br><br>+Comparable to the performance of the 3 individual experts (96%, 94%, 92%), and significantly higher than the mean performance of 31 nonexperts (81%) |
| | | | | | | - Used manually segmented images with a tracing algorithm to avoid the possible noise and bias that might come from an automated segmentation algorithm<br>-Low clinical applicability |
| Rani et al, 2016 [70] | Back Propagation Neural Networks | To classify ROP | | 64 RGB images of these stages have been taken, captured by RetCam with 120 degrees field of view and size of 640 x 480 pixels. | 90.6% (accuracy) | |
| | | | | | | -No clinical information<br>-Required better segmentation<br>-Clinical adaptation |
| Karayiannis et al, 2006 [71] | Artificial Neural Networks (ANN) | To aim at the development of a seizure-detection system by training neural networks with quantitative motion information extracted from short video segments of neonatal seizures of the myoclonic and focal clonic types and random infant movements | 54 patients | 240 video segments ( Each of the training and testing sets contained 120 video segments (40 segments of myoclonic seizures, 40 segments of focal clonic seizures, and 40 segments of random movements | 96.8% (sensitivity)<br>97.8% (specificity) | +Video analysis |
| | | | | | | - Not be capable of detecting neonatal seizures with subtle clinical manifestations (Subclinical seizures) or neonatal seizures with no clinical manifestations (electrical-only seizures<br>-Not include EEG analysis<br>-Small sample size<br>-No additional clinical information |

**Table 2:** ML based (non-DL) studies in neonatology using non-imaging data for diagnosis

| Study | Approach | Purpose | Dataset | Type of data | Performance | Pros( +) |
|---|---|---|---|---|---|---|
| | | | | | | Cons (-) |
| Reed et al , 1996[72] | Recognition-based reasoning | Diagnosis of congenital heart defects | 53 patients | Patient history, physical exam, blood tests, cardiac auscultation, X-ray, and EKG data | | + Useful in multiple defects<br>-Small sample size<br>-Not real AI implementation |
| Aucouturier et al, 2011[73] | Hidden Markov model architecture (SVM, GMM) | To identify expiratory and inspiration phases from the audio recording of human baby cries | 14 infants, spanning four vocalization contexts in their first 12 months | Voice record- | 86%-95% (accuracy) | + Quantify expiration duration, count the crying rate, and other time-related characteristics of baby crying for screening, diagnosis, and research purposes over large populations of infants<br><br>+Preliminary result<br>-More data needed<br>-No clinical explanation<br>-Small sample size<br>-Required preprocessing |
| Cano Ortiz et al, 2004[74] | Artificial neural networks (ANN) | To detect CNS diseases in infant cry | 35 neonates, nineteen healthy cases and sixteen sick neonates | Voice record (187 patterns) | 85% (accuracy) | +Preliminary result<br>-More data needed for correct classification for |
| Hsu et al, 2010[75] | Support Vector Machine (SVM) Service-Oriented Architecture (SOA | To diagnose Methylmalonic Acidemia (MMA) | 360 newborn samples | Metabolic substances data collected from tandem mass spectrometry (MS/MS) | 96.8% (accuracy) | +Better sensitivity than classical screening methods<br>-Small sample size<br>- SVM pilot stage education not integrated |
| Baumgartner et al, 2004 [76] | Logistic regression analysis (LRA) Support vector machines (SVM) Artificial neural networks (ANN) Decision trees (DT) k-nearest neighbor classifier (k-NN) | Focusing on phenylketonuria (PKU), medium chain acyl-CoA dehydrogenase deficiency (MCADD | During the Bavarian newborn screening program all newborns | Metabolic substances data collected from tandem mass spectrometry (MS/MS) | 99.5% (accuracy) | +ML techniques, LRA (as discussed above), SVM and ANN, delivered results of high predictive power when running on full as well as on reduced feature dimensionality.<br>- Lacking direct interpretation of |
| | | | | | | the knowledge representation |
| Chen et al, 2013[77] | Support vector machine (SVM) | To diagnose phenylketonuria (PKU), hypermethioninemia, and 3-methylcrotonyl-CoA-carboxylase (3-MCC) deficiency | 347,312 infants (220 metabolic disease suspect) | Newborn dried blood samples | 99.9% (accuracy) 99.9% (accuracy) 99.9% (accuracy) | +Reduced false positive cases<br>- The feature selection strategies did not include the total features for establishing either the manifested features or total combinations |
| Temko et al, 2011[78] | Support Vector Machine (SVM) classifier leave-one-out (LOO) cross-validation method. | To measure system performance for the task of neonatal seizure detection using EEG | 17 newborns system is validated on a large clinical dataset of 267 h All seizures were annotated independently by 2 experienced neonatal electroencephalographers using video EEG | EEG data | 89% (AUC) | + SVM-based seizure detection system can greatly assist clinical staff, in a neonatal intensive care unit, to interpret the EEG<br>- No clinical variable<br>- Datasets for neonatal seizure detection are quite difficult to obtain and never too large division results in a potentially large bias. |
| Temko et al, 2012 [79] | SVM | To use recent advances in the clinical understanding of the temporal evolution of seizure burden in neonates with hypoxic ischemic encephalopathy to improve the performance of automated detection algorithms. | 17 HIE patients | 816.7 hours EEG recordings of infants with HIE | 96.7% (AUC) | +Improved seizure detection |
| Temko et al, 2013[80] | Support Vector Machine (SVM) classifier leave-one-out (LOO) cross-validation method | Robustness of Temko 2011[78] | Trained in 38 term neonates<br><br>Tested in 51 neonates | Trained in 479 hours EEG recording Tested in 2540 hours | 96.1% (AUC)<br><br>Correct detection of seizure burden 70% | -Small sample size<br>-No clinical information |
| Stevenson et al, 2013[81] | Multiclass linear classifier | Automatically grading one hour EEG epoch | 54 full term neonates | One-hour-long EEG recordings | 77.8% (accuracy) | +Involvement of clinical expert<br>+Method explained in a detailed way<br>Retrospective design |
| Ahmed et al, 2016[82] | -Gaussian mixture model.<br>-Universal Background Model (UBM)<br>-SVM | An automated system for grading hypoxic–ischemic encephalopathy (HIE) severity using EEG is presented | 54 full term neonates (same dataset as Stevenson et al 2013) | One-hour-long EEG recordings | 87% (accuracy) | +Provide significant assistance to healthcare professionals in assessing the severity of HIE<br>+Some brief temporal activities (spikes, sharp waves and certain spatial characteristics such as asynchrony and asymmetry) which are not detected by system<br>-Retrospective design |
| Mathieson et al, 2016[83] | Robusted Support Vector Machine (SVM) classifier leave-one-out (LOO) cross-validation method [80] | Validation of Temko 2013[80] | 70 babies from 2 centers<br><br>35 Seizure<br>35 Non Seizure | | Seizure detection Algorithm thresholds is clinically acceptable range Detection rates 52.5%-75% | +Clinical information and Cohen score were added<br>+First Multi center study<br>-Retrospective design |
| Mathieson et al, 2016[84] | Support Vector Machine (SVM) classifier leave-one-out (LOO) cross-validation method. [78] | Analysis of Seizure detection Algorithm and characterization of false negative seizures | 20 babies(10 seizure -10 non seizure)<br><br>( 20 of 70 babies) [83] | | Seizure detections were evaluated the sensitivity threshold | +Clinical information and Cohen score were added<br>+Seizure features were analyzed<br>-Retrospective design |
| Yassin et al, 2017[85] | Locally linear embedding (LLE) | Explore autoencoders to perform diagnosis of infant asphyxia from infant cry | | One-second segmentation was then performed producing 600 segmented signals, from which 284 were normal cries while 316 were asphyxiated cries | 100% (accuracy) | +600 MFCC features of normal and non-asphyxiated newborns<br>-No clinical information |
| Li et al, 2011[86] | Fuzzy backpropagation neural networks | To establish an early diagnostic system for hypoxic ischemic encephalopathy (HIE) in newborns | 140 cases (90 patients and 50 control) | The medical records of newborns with HIE | The correct recognition rate was 100% for the training samples, and the correct recognition rate was 95% for the test samples, indicating a misdiagnosis rate of 5%. | +High accuracy in the early diagnosis of HIE<br>-Small sample size |
| Zernikow et al, 1998[87] | ANN | To detect early and accurately the occurrence of severe IVH in an individual patient | 890 preterm neonates (50%, 50%) Validation and training | EHR | 93.5% (AUC) | +Observational study<br>+Skipped variables during training of ANN<br>-No image |
| Ferreira et al, 2012[88] | Decision trees and neural networks | Employing data analysis methods to the problem of identifying neonatal jaundice | 227 healthy newborns | 70 variables were collected and analyzed | 89% (accuracy)<br>84% (AUC) | + Predicting subsequent hyperbilirubinemia with high accuracy<br>+ Data mining has the potential to assist in clinical decision - making, thus contributing to a more accurate diagnosis of neonatal jaundice<br>-Not included all factors contributing to hyperbilirubinemia |
| Porcelli et al, 2010[89] | Artificial neural network (ANN) | To compare the accuracy of birth weight–based weight curves with weight curves created from individual patient records | 92 ELBW infants | Postnatal EHR | The neural network maintained the highest accuracy during the first postnatal month compared with the static and multiple regression methods | +ANN-generated weight curves more closely approximated ELBW infant weight curves, and, using the present electronic health record systems, may produce weight curves better reflective of the patient's status |
| Mueller et al, 2004[90] | Artificial neural network (ANN) and a multivariate logistic regression model (MLR). | To compare extubation failure in NICU | 183 infants (training (130) / validation(53)) | EHR, 51 potentially predictive variables for extubation decisions | 87% (AUC) | +Identification of numerous variables considered relevant for the decision whether to extubate a mechanically ventilated premature infant with respiratory distress syndrome<br>-Small sample size<br>-2-hour prior extubation took into consideration<br>-Longer duration should be encountered |
| Precup et al, 2012[91] | Support Vector Machines (SVM) | To determine the optimal time for extubation that will minimize the duration of MV and maximize the chances of success | 56 infants; 44 successfully extubated and 12 required re-intubation | Respiratory and ECG signals 3,000 samples of the AUC features for each baby | 83.2% (failure class-accuracy)<br><br>73.6% ( success class-accuracy) | +Prospective<br>-Small sample size<br>-Overfitting |
| Hatzakis et al, 2002[92] | Fuzzy Logic Controller | To develop modularized components for weaning newborns with lung disease | 10 infants with severe cyanotic congenital heart disease following surgical procedures requiring intra-operative cardiac bypass support | Through respiratory frequency (RR); tidal volume (VT); minute ventilation (VE); gas diffusion (PaO2, PaCO2, P(A-a)02 and pH); muscle effort parameters of oxygen saturation (SaO2) and heart rate (HR) | -No evaluation metrics | +More intelligent systems<br>-Surrogate markers relevant to virus, drug, host, and mechanical ventilation interactions will have to be considered<br>-Retrospective |
| Dai et al, 2021[93] | ML | To determine the significance of genetic variables in BPD risk prediction early and accurately | 131 BPD infants and 114 infants without BPD | Clinical Exome sequencing(Thirty and 21 genes were included in BPD–RGS and sBPD) | 90.7% (sBPD-AUC)<br><br>91.5% (BPD-AUC) | + Conducted a case–control analysis based on a prospective preterm cohort<br>+Genetic information contributes to susceptibility to BPD<br>+Data available<br>- A single-center design leads to missing data and unavoidable biases in identifying and recruiting participants |
| Tsien et al, 2000 [94] | C4.5 Decision tree system (artefact annotation by experts) | To detect artifact pattern across multiple physiologic data signals | Data from bedside monitors in the neonatal ICU | 200 h of four-signal data(ECG,HR,BP,$CO_2$ | 99.9% ($O_2$-AUC)<br><br>93.3% ($CO_2$ -AUC)<br><br>89.4% (BP-AUC)<br><br>92.8% (HR-AUC) | + Annotations would be created prospectively with adequate details for understanding any surrounding clinical conditions occurring during alarms<br>- The methodology employed for data annotation<br>-Retrospective design<br>-Not confirmed with real clinical situations data may not |
| | | | | | | -Capture short-lived artifacts and thus these models would not be effectively designed to detect such artifacts in a prospective setting |
| Koolen et al, 2017[95] | SVM | To develop an automated neonatal sleep state classification approach based on EEG that can be employed over a wide age range | 231 EEG recordings from 67 infants between 24 and 45 weeks of postmenstrual age. Ten-minute epochs of 8 channel polysomnography (N = 323) from active and quiet sleep were used as a training dataset. | A set of 57 EEG features | 85% (accuracy) | + A robust EEG-based sleep state classifier was developed + The visualization of sleep state in preterm infants which can assist clinical management in the neonatal intensive care unit +Clinical variables<br>-No integration of physiological variables -Need of longer records |
| Mohseni et al, 2006[96] | Artificial neural network (ANN) | To detect EEG rhythmic pattern detection | 4 infants | 2-hour EEG record | 72.4% (sensitivity)<br><br>93.2% (specificity) | +Uses very short (0.4 second) segment of the data in compared to the other methods (10 seconds), + Detect seizure sooner and more accurately<br>-Small sample size -No clinical information |
| Simayijiang et al, 2013 [97] | Random Forest (RF) | To analyze the features of EEG activity bursts for predicting outcome in extremely preterm infants. | 14 extremely preterm infants Eight infants had good outcome and six had poor outcome, defined as neurodevelopmental impairment according to psychological testing and neurological examination at two years age | One-channel EEG recordings during the first three postnatal days of 14 extremely preterm infants | 71.4% (accuracy) | + Each burst six features were extracted and random forest techniques<br>-Small sample size |
| Ansari et al, 2015 [98] | SVM | To reduce EEG artifacts in NICU | 17 neonates (for training) | 27 hours recording EEG | | + Reduced false alarm rate |
| | | | 18 neonates for testing | polygraphy (ECG, EMG, EOG, abdominal respiratory movement signal | False alarm rate drops 42% | -Small sample size<br>-Not fully online |
| Matic et al, 2016 [99] | Least-squares support vector machine (LS-SVM) classifiers<br><br>low-amplitude temporal profile (LTP). | To develop an automated algorithm to quantify background electroencephalography (EEG) dynamics in term neonates with hypoxic ischemic encephalopathy | 53 neonates | The recordings were started 2–48 (median 19) hours postpartum, using a set of 17 EEG electrodes, whereas in some patients, a reduced set of 13 electrodes was used | 91% (AUC)<br>94% (AUC)<br>94% (AUC)<br>97% (AUC) | +The first study that used an automated method to study EEGs over long monitoring hours and to accurately detect milder EEG discontinuities<br>+ Necessary to perform further multicenter validation studies with even larger datasets and characterizing patterns of brain injury on MRI and clinical outcome<br>- The number of misclassifications was rather high as compared to the EEG expert |
| Navarro et al, 2017 [100] | kNN, SVM and LR | To detect EEG burst in preterm infants | Trained 14 very preterm infants Testing in 21 infants | EEG recording | 84% (accuracy) | + New functionality to current bedside monitors,<br>+ Integrating wearable devices or EEG portable headsets) to follow up maturation in preterm infants after hospital discharge |
| Ahmed et al, 2017 [101] | Gaussian dynamic time warping<br>SVM<br>Fusion | To improve the detection of short seizure events | 17 neonates | EEG recording ( 261 h of EEG) | 71.9% (AUC)<br>69.8% (AUC)<br>75.2% (AUC) | +Achieving a 12% improvement in the detection of short seizure events over the |
| | | | | | | static RBF kernel based system<br>-Better post processing methods<br>-Small sample size |
| Thomas, et al, 2008 [102] | Basic Gradient Descent (BGD)<br>Least Mean Squares (LMS)<br>Newton Least Mean Squares (NLMS) | To alert NICU staff ongoing seizures and detect neonatal seizures | 17 full term neonates | EEG recording | 77% (Global classifier-AUC)<br>80% (BGD-AUC)<br>79% (LMS-AUC)<br>80% (NLMS-AUC) | + The adapted classifiers outperform the global classifier in both sensitivity and specificity leading to a large increase in accuracy<br>- Local training data is not representative of the patient's entire EEG record |
| Schetinin et al, 2004 [103] | Artificial Neural Networks (ANN)<br><br>(GMDH :Group Method of Data Handling)<br><br>(DT:Decision Tree)<br><br>FNN: Feedforward Neural Network<br><br>PNN:Polynomial Neural Network (<br><br>Combined (PNN&DT) | To detect artifacts in clinical EEG of sleeping newborns | 42 neonates | 40 EEG records<br><br>20 records containing 17 094 segments were randomly selected for training<br><br>20 records containing 21 250 segments were used for testing | 69.8% (DT-accuracy)<br><br>70.7% (FNN-accuracy)<br><br>73.2% (GMDH-accuracy)<br><br>73.2% (PNN-accuracy)<br><br>73.5% (PNN&DT) | + Keep the classification error done<br>- Not included other signal data (EMG, EOG) |
| Na et al, 2021 [104] | Multiple Logistic Regression | Compare the performance of AI analysis with that of conventional analysis to identify risk factors associated with symptomatic PDA (sPDA) in very low birth weight infants | 10390 Very low birth weight infant | 47 perinatal risk factors | 77% (75%-79%) (accuracy)<br><br>82% (80%-84%) (AUC) | +First to use AI to predict sPDA and sPDA therapy and to analyze the main risk factors for sPDA using large-scale cohort data comprising only electronic records<br>-Low accuracy<br>-Non image dataset |
| Gómez-Quintana et al, 2021 [105] | XGBoost | Developing an objective clinical decision support tool based on | 265 infants | Phonocardiogram | 88% (AUC) | +PDA diagnosis with phonocardiogram |
| | | ML to facilitate differentiation of sounds with signatures of Patent Ductus Arteriosus (PDA)/CHDs, in clinical settings | | | | -Worst performance in early days of life which is more important for diagnosis<br>-Low prediction rate with ML<br>. |
| Sentner et al, 2022 [106] | Logistic regression, decision tree, and random forest | To develop an automated algorithm based on routinely measured vital parameters to classify sleep-wake states of preterm infants in real-time at the bedside. | 37 infants (PMA:31.1 ± 1.5 weeks<br><br>9 infants(PMA 30.9 ± 1.3) validation | Sleep-wake state observations were obtained in 1-minute epochs using a behavioral scale developed in-house while vital signs (HR, RR, $SO_2$ were recorded simultaneously) | 80% (AUC)<br>77% (AUC) | +Real-time sleep staging algorithm was developed for the first time for preterm infants<br>+Adapt bedside clinical work based on infants" sleep-wake states, potentially promoting the early brain development and well-being of preterm infants<br>+without EEG signals, noninvasive tool<br>+Observational study<br>- Small sample size<br>- No additional clinical information |
| Pavel et al, 2020[107] | ANSeR Software System<br><br>SVM<br>GMM<br>Universal Background Model (UBM), | To detect neonatal seizure with algorithm | 128 neonates in algorithm group<br><br>130 neonates in non algorithm group | 2 -100 hours EEG recording for each neonate | Specificity<br>Sensitivity<br>False Alarm Rate were calculated.<br><br>AUC and accuracy were not calculated.<br><br>Seizures detected by algorithm<br><br>No difference between the algorithm and non-algorithm group specificity, sensitivity | + The first randomized, multicenter clinical investigation to assess the clinical impact of a machine-learning algorithm in real time on neonatal seizure recognition in a clinical setting<br><br>-The authors mentioned the algorithm [78,80,83]but not defined detailed way |
| Mooney et al, 2021[108] | Random Forest | Secondary analysis of Validation of Biomarkers in HIE ( BiHiVE study) | 53000 birth screened 409 infants were included<br><br>129 infants with HIE | 154 clinical variables<br><br>Blood gas analysis<br><br>APGAR | Three model were used for analysis<br><br>Best evaluation metrics Accuracy: 94% Specificity: 92% Sensitivity: 100% | + Classification with ML<br><br>+ Secondary analysis of prior prospective trial<br><br>-Not a prospective design |

**Table 3:** ML based (non-DL) studies in neonatology using imaging data for prediction.

| Study | Approach | Purpose | Dataset | Type of data | Performance | Pros(+) / Cons(-) |
|---|---|---|---|---|---|---|
| Vassar et al, 2020 [109] | Multivariate models with leave-one-out cross-validation and exhaustive feature selection | Very premature infants' structural brain MRI and white matter microstructure as evaluated by diffusion tensor imaging (DTI) in the near term and their impact on early language development | 102 infants | Brain MRI and DTI + ( Bayley Scales of Infant-Toddler Development-III at 18 to 22 months) | 50.2% (language composite score -AUC) 61.7% (expressive language subscore-AUC) 32.2% (receptive language subscore-AUC) | + Preterm babies at risk for language impairment may be identified using multivariate models of near-term structural MRI and white matter microstructure on DTI, allowing for early intervention<br>- Demographic data is not included -Cross validation? -Small sample size |
| Schadl et al, 2018 [110] | -Linear models with exhaustive feature selection and leave-one-out cross-validation | To predict neurodevelopment in preterm children in near term MRI and DTI | 66 preterm infants | Brain MRI and DTI 51 WM regions (48 bilateral regions, 3 regions of corpus callosum)<br><br>Bayley Scales of Infant-Toddler Development, 3rd-edition (BSID-III) at 18–22 months. | 100% (AUC, cognitive impairment) 91% (AUC, motor impairment | - Using structural brain MRI findings of WMA score, lower accuracy -Small cohort -DTI has better implementation and interpretation |
| Wee et al, 2017 [111] | SVM and canonical correlation analysis (CCA) | To examine heterogeneity of neonatal brain network and its prediction to | 120 neonates | 1.5-Tesla DW MRI Scans<br><br>Diffusion tensor imaging (DTI) tractography | 89.4% (accuracy) | +Neural organization established during fetal development could to some extent predict individual differences in |
| | | child behaviors at 24 and 48 months of age | | Child Behavior Checklist (CBCL) at 24 and 48 months of age. | | behavioral emotional problems in early childhood<br>-Small sample size |

**Table 4:** ML based (non-DL) studies in neonatology using non-imaging data for prediction.

| Reference | Approach | Purpose | Dataset | Type of data | Performance | Pros( +)<br>Cons (-) |
|---|---|---|---|---|---|---|
| Soleimani et al, 2012 [112] | Multilayer perceptron (MLP) (ANN) | Predict developmental disorder | 6150 infants' | Infant Neurological International Battery (INFANIB) and prenatal factors | 79% (AUC) | +Neural network ability includes quantitative and qualitative data<br>-Relying on preexisting data<br>-Missing important topics<br>-Small sample size |
| Zernikow et al, 1998 [113] | ANN | To predict the individual neonatal mortality risk | 890 preterm neonates | Clinical records | 95% (AUC) | +ANN predict mortality accurately<br>- Its high rate of prediction failure |
| Ji et al, 2014 [114] | Generalized linear mixed-effects models | To develop the NEC diagnostic and prognostic models | 520 infants | Clinical variables | 84%-85% (AUC) | + Prediction of NEC and risk stratification.<br>- Non image data |
| Young et al, 2012 [115] | Multilayer perceptron (MLP) ANN | To forecasting the sound loads in NICUs | 72 individual data | Voice record- | | + Prediction of noise levels<br>- Limited only to time and noise level |
| Nascimento LFC et al., 2002 [116] | A fuzzy linguistic model | To estimate the possibility of neonatal mortality. | 58 neonatal deaths in 1,351 records. | EHR | It depends on the GA, APGAR score and BW<br><br>90% (accuracy) | + Not to compare this model with other predictive models because the fuzzy model does not use blood analyses and current models such as PRISM, SNAP or CRIB do not use the fuzzy variables<br>- No change over the time |
| Reis et al, 2004 [117] | Fuzzy composition | Determine if more intensive neonatal resuscitation procedures will be required during labor and delivery | Nine neonatologists facing which a degree of association with the risk of occurrence of perinatal asphyxia | 61 antenatal and intrapartum clinical situations | 93% (AUC) | + Maternal medical, obstetric and neonatal characteristics to the clinical conditions of the newborn, providing a risk measurement of need of advanced neonatal resuscitation |
| | | | | | | measures<br>- Implement a supplemental system to help health care workers in making perinatal care decisions.<br>- Eighteen of the factors studied were not tested by experimental analysis, for which testing in a multicenter study or over a very long period of time in a prospective study would be probably needed<br>-No image |
| Jalali et al, 2018 [118] | SVM | To predict the development of PVL by analyzing vital sign and laboratory data received from neonates shortly following heart surgery | 71 neonates( including HLHS and TGA) | Physiological and clinical data Up to 12 h after cardiac surgery | 88% (AUC) | + Might be used as an early prediction tool<br>- Retrospective observational study<br>- Other variables did not collected which precipitated the PVL |
| Ambalavanan et al, 2000 [119] | ANN | To predict adverse neurodevelopmental outcome in ELBW | 218 neonates 144 for training 74 for test set | Clinical variables and Bayley scores at 18 months | 62% (Major handicapped-AUC) | +Neural network is more sensitive detection individual mortality<br>-Short follow up<br>- Underperformance of neural network |
| Saria et al, 2010 [120] | Bayesian modeling paradigm<br><br>Leave one out algorithm | To develop morbidity prediction tool | To identify infants who are at risk of short- and long-term morbidity in advance | Electronically collected physiological data from the first 3 hours of life in preterm newborns (<34 weeks gestation, birth weight <2000 gram) of 138 infants | 91.9% (AUC-predicting high morbidity) | + Physiological variables, notably short-term variability in respiratory and heart rates, contributed more to morbidity prediction than invasive laboratory tests. |
| Saadah et al, 2014 [121] | ANN | To identify subgroups of premature infants who may benefit from palivizumab prophylaxis during nosocomial outbreaks of respiratory | 176 infants 31 (17.6%) received palivizumab during the outbreaks | EHR | In male infants whose birth weight was less than 0.7 kg and who had hemodynamically significant congenital heart disease. | - Retrospective analysis using an AI model<br>-No external validation<br>- Low generalizability<br>- Small sample size |
| | | syncytial virus (RSV) infection | | | | |
| Mikhno et al, 2012 [122] | Logistic Regression Analysis | Developed a prediction algorithm to distinguish patients whose extubation attempt was successful from those that had EF | 179 neonates | EHR 57 candidate features Retrospective data from the MIMIC-II database | 87.1% (AUC) | + A new model for EF prediction developed with logistic regression, and six variables were discovered through ML techniques<br>- 2 hour prior extubation took into consideration -longer duration should be encountered |
| Gomez et al, 2019[123] | AdaBoost Bagged Classification Trees (BCT) Random Forest(RF) Logistic Regression (LR) SVM | To predict sepsis in term neonates within 48 hours of life monitoring heart rate variability(HRV) and EHR | 79 newborns<br><br>15 were diagnosed with sepsis | 4 EHR variables and HRV variables .<br><br>HRV variables were analyzed with the ML methods | 94.3% (AUC) AdaBoost<br><br>88.8% (AUC) Bagged Classification Trees<br><br>Lowest AUC 64% ( k-NN) | + Noninvasive methods for sepsis prediction<br>- Small sample size<br>- Need an extra software for HRV analysis<br>- Not included EHR into ML analysis<br>- No Adequate Clinical Information |
| Verder et al, 2020 [124] | Support vector machine (SVM) | To develop a fast bedside test for prediction and early targeted intervention of bronchopulmonary dysplasia (BPD) to improve the outcome | 61 very preterm infants were included in the study | Spectral pattern analysis of gastric aspirate combined with specific clinical data points | Sensitivity: 88%<br><br>Specificity: 91% | + Multicenter non-interventional diagnostic cohort Study<br>+ Early prediction and targeted intervention of BPD have the potential to improve the outcome<br>+First algorithm developed by AI to predict BPD shortly after birth with high sensitivity and specificity.<br>-Small sample size |
| Ochab et al, 2015 [125] | | To predict BPD in LBW infant | 109 neonates | EHR (14 risk factors) | 83.2% (accuracy) | + Decision support system |
| | SVM and logistic regression | | | | | - Small sample size<br>- Few clinical variables<br>- Low accuracy with SVM<br>- A single-center design leads to missing data and unavoidable biases in identifying and recruiting participants |
| Townsend et al, 2008 [126] | ANN | To predict events in the NICU | Data collected by the CNN between January 1996 and October 1997 contains data from 17 NICUs | 27 clinical variables | 85% (AUC) | + Modeling life-threatening complications will be combined with a case-presentation tool to provide physicians with a patient's estimated risk for several important outcomes<br>+ Annotations would be created prospectively with adequate details for understanding any surrounding clinical conditions occurring during alarms<br>- The methodology employed for data annotation<br>-Retrospective design<br>- Not confirmed with real clinical situations<br>- Data may not capture short-lived artifacts and thus these models would not be effectively designed to detect such artifacts in a prospective setting |
| Ambalavanan et al, 2005 [127] | ANN and logistic regression | To predict death of ELBW infant | 8608 ELBW infants | 28 clinical variables | 84% (AUC)<br>85% (AUC) | + The difficulties of predicting death should be acknowledged in discussions with families and caregivers about decisions regarding initiation or continuation of care |
| | | | | | | -<br>Chorioamnionitis, timing of prenatal steroid therapy, fetal biophysical profile, and resuscitation variables such as parental or physician wishes regarding resuscitation) could not be evaluated because they were not part of the data collected. |
| Bahado-Singh et al, 2022 [128] | Random forest (RF), support vector machine (SVM), linear discriminant analysis (LDA), prediction analysis for microarrays (PAM), and generalized linear model (GLM) | Prediction of coarctation in neonates | Genome-wide DNA methylation analysis of newborn blood DNA | 24 patients<br>16 controls | 97% (80%–100%) (AUC) | + AI in epigenomics<br>+ Accurate prediction of CoA ‘<br>-Small dataset<br>-Not included other CHD |
| Bartz-Kurycki et al, 2018 [129] | Random forest classification (RFC), and a hybrid model (combination of clinical knowledge and significant variables from RF) | To predict neonatal surgical site infections (SSI) | 16,842 neonates | EHR | 68% (AUC) | +Large dataset<br>+Important neonatal outcome<br>- Retrospective study<br>- Bias in missing data |
| Do et al, 2022 [130] | Artificial neural network (ANN), random forest (RF), and support vector machine (SVM) | To predict mortality of very low birth weight infants (VLBWI) | 7472 VLBWI data from Korean neonatal network | EHR | 84.5% (81.5%-87.5%)(ANN-AUC)<br><br>82.6%(79.5%-85.8%) (RF-AUC)<br><br>63.1% (57.8%-68.3%). SVM-AUC | + VLBWI mortality prediction using ML methods would produce the same prediction rate as the standard statistical LR approach and may be appropriate for predicting mortality studies utilizing ML confront a high risk of selection bias. |
| | | | | | | - Low prediction rate with ML |
| Podda et al, 2018 [131] | ANN | Development of the Preterm Infants Survival Assessment (PISA) predictor | Between 2008 and 2014, 23747 neonates (<30 weeks gestational age or <1501 g birth weight were recruited Italian Neonatal Network | 12 easily collected perinatal variables | 91.3% (AUC)<br>77.9% (AUC)<br>82.8% (AUC)<br>88.6% (AUC) | + NN had a slightly better discrimination than logistic regression<br>- Like all other model-based methods, is still too imprecise to be used for predicting an individual infant's outcome<br>- Retrospective design<br>- Lack of variables |
| Turova et al, 2020 [132] | Random Forest | To predict intraventricular hemorrhage in 23-30 weeks of GA infants | 229 infants | Clinical variables and cerebral blood flow (extracted from mathematical calculation) were used<br><br>10 fold validation | 86%-93% ( AUC)<br><br>Vary on the extracted features in and feature weight in the model | + Good accuracy<br>- Retrospective<br>- Gender distribution was not standardized between the groups<br><br>- Not corresponding lab value according to the IVH time |
| Cabrera-Quiros et al, 2021[133] | Logistic regressor, naive Bayes, and nearest mean classifier | Prediction of late-onset sepsis (starting after the third day of life) in preterm babies based on various patient monitoring data 24 hours before onset | 32 premature infants with sepsis and 32 age-matched control patients | Heart rate variability, respiration, and body motion, differences between late-onset sepsis and Control group were visible up to 5 hours preceding the cultures, resuscitation, and antibiotics started here(CRASH) point | Combination of all features showed a mean accuracy 79% and mean precision rate 82%<br>3 hours before the onset of sepsis<br><br>Naive Bayes accuracy :71%<br>Nearest Mean :70% | + Monitoring of vital parameters could be predicted late onset sepsis up to 5 hours.<br>- Small sample size<br>- Retrospective<br>- Gestational age, postnatal age, sepsis and culture |
| Reed et al,2021 [134] | Comparison least absolute shrinkage and selection operator (LASSO) and random forest (RF) to expert-opinion driven logistic regression modelling | Prediction of 30-day unplanned rehospitalization of preterm babies | 5567 live-born babies and 3841 were included to the study<br><br>Data derived exclusively from The population-based prospective cohort study of French preterm babies, EPIPAGE 2. | The logistic regression model comprised 10 predictors, selected by expert clinicians, while the LASSO and random forest included 75 predictors | 65% (AUC) RF<br><br>59% (AUC) LASSO<br><br>57% (AUC) LR | +The first comparison of different modelling methods for predicting early rehospitalization<br><br>+Large cohort with data variation<br><br>-No accurate evaluation of rehospitalization causes<br><br>-Data collection after discharge based on survey filled by mothers<br><br>-9% of babies were rehospitalized |
| Khursid et al, 2021[135] | K-nearest neighbor, random forest, artificial neural network, stacking neural network ensemble | To predict, on days 1, 7, and 14 of admission to neonatal intensive care, the composite outcome of BPD/death prior to discharge. | <33 weeks GA cohort (n = 9006)<br><br>And < 29 weeks GA were included | For each set of models (Days 1, 7, 14), stratified random sampling. 80% of used were training. 20% of used were test set. 10-fold cross validation for test dataset | 81%-86% (AUC) for , 33 weeks<br><br>70-79% (AUC) for , 29 weeks | + Large dataset<br><br>- Not having good performance scores<br><br>- No data sharing<br><br>-Not included important predictors ($FiO_2$ and presence of PDA before 7th days ) |
| Moreira et al, 2022 [136] | Logistic regression and Random Forest | To develop an early prediction model of neonatal death on extremely low gestational age(ELGA ) infants | < 28 weeks Swedish Neonatal Quality Registry 2011-May 2021<br><br>3752 live born ELGA infants | Birth weight, Apgar score at 5 min, gestational age were selected as features and new model (BAG) designed to predict mortality | 76.9%(AUC)<br><br>Validation cohort 68.9% (AUC) | +Model development cohort and validation cohort included<br><br>+ BAG model had better AUC than individual birthweight and gestational age model.<br><br>+Code is available<br><br>+ Online calculator is available<br><br>- BAG model does not include clinical variables and clinical practice. Birthweight and gestational age could not be changed. Only Apgar scores could be changed. |
| | | | | | | |
| Hsu et al,2020[137] | RF<br>KNN<br>ANN<br>XGBoost<br>Elastic-net | To predict mortality of neonates when they were on mechanical intubation | 1734 neonates<br>70% training<br>30% test | Mortality scores<br>Patient demographics<br>Lab results<br>Blood gas analysis<br>Respirator parameters<br>Cardiac inotrop agents from onset of respiratory failure to 48 hours | 93.9% (AUC)<br>RF<br><br>has achieved the highest prediction of mortality | +Employed several ML and statistics<br><br>+Explained the feature analysis and importance into analysis<br><br>- Two center study<br><br>-Algorithmic bias<br><br>-Inability to real time prediction |
| Stocker et al, 2022 [138] | RF | To predict blood culture test positivity according to the all variables, all variables without biomarkers, only biomarkers, only risk factors, and only clinical signs | 1710 neonates from 17 centers<br><br>Secondary analysis of NeoPInS data | Biomarkers(4 variables)<br><br>Risk factors (4 variables)<br><br>Clinical signs(6 variables)<br><br>Other variables(14)<br><br>All variables (28)<br><br>They included to RF analysis to predict culture positive early onset sepsis | Only biomarkers<br>73.3% (AUC)<br><br>All variables<br>83.4% (AUC)<br><br>Biomarkers are the most important contributor | +CRP and WBC are the most important variables in the model<br>+ Decrease the overtreatment<br>+Multi center data<br><br>- Overfitting of the model due to the discrepancy with currently known clinical practice<br>- Seemed not evaluated the clinical signs and risk factors which are really important in daily practice |

**Table 6:** DL based studies in neonatology using imaging and non-imaging data for diagnosis.

| Study | Approach | Purpose | Dataset | Type of data ( Image/Non Image) | Performance | Pros(+)<br><br>Cons(-) |
|---|---|---|---|---|---|---|
| Hauptmann et al, 2019 [139] | 3D (2D plus time) CNN architecture | Ability of CNNs to reconstruct highly accelerated radial real-time data in patients with congenital heart disease | 250 CHD patients. | Cardiovascular MRI with cine images | | +Potential use of a CNN for reconstruction real time radial data |
| Lei et al, 2022 [140] | MobileNet-V2 CNN | Detect PDA with AI | 300 patients | Echocardiography | 88% (AUC) | + Diagnosis of PDA with AI |
| | | | 461 echocardiograms | | | - Does not detect the position of PDA |
| Ornek et al, 2021 [141] | VGG16 (CNN) | To focus on dedicated regions to monitor the neonates and decides the health status of the neonates (healthy/ unhealthy) | 38 neonates | 3800 Neonatal thermograms | 95% (accuracy) | + Known with this study how VGG16 decides on neonatal thermograms<br>- Without clinical explanation |
| Ervural et al, 2021[142] | Data Augmentation and CNN | Detect health status of neonates | 44 neonates | 880 images Neonatal thermograms | 62,2% to 94,5% (accuracy) | + Significant results with data augmentation<br>- Less clinically applicable<br>- Small dataset |
| Ervural et al, 2021[143] | Deep siamese neural network(D-SNN) | Prediagnosis to experts in disease detection in neonates | 67 neonates, | 1340 images Neonatal thermograms | 99.4% (infection diseases accuracy in 96.4% (oesophageal atresia accuracy), 97.4% (in intestinal atresia-accuracy,<br><br>94.02% (necrotising enterocolitis accuracy) | + D-SNN is effective in the classification of neonatal diseases with limited data<br>- Small sample size |
| Ceschin et al, 2018 [144] | 3DCNNs | Automated classification of brain dysmaturation from neonatal MRI in CHD | 90 term-born neonates with congenital heart disease and 40 term-born healthy controls | 3 T brain MRI | 98.5% (accuracy) | + 3D CNN on small sample size, showing excellent performance using cross-validation for assessment of subcortical neonatal brain dysmaturity<br>+ Cerebellar dysplasia in CHD patients<br>-Small sample size |
| Ding et al, 2020 [145] | HyperDense-Net and LiviaNET | Neonatal brain segmentation | 40 neonates<br>24 for training<br>16 for experiment | 3T Brain MRI T1 and T2 | 94% 95%/ 92% (Dice Score)<br><br>90% /90% / 88% (Dice Score) | +Both neural networks can segment neonatal brains, achieving previously reported performance |
| | | | | | | -Small sample size |
| Liu et al, 2020 [146] | Graph Convolutional Network (GCN) | Brain age prediction from MRI | 137 preterm | 1.5-Tesla MRI<br>+<br><br>Bayley-III Scales of Toddler Development at 3 years | Show the GCN's superior prediction accuracy compared to state-of-the-art methods | + The first study that uses GCN on brain surface meshes to predict neonatal brain age, to predict individual brain age by incorporating GCN-based DL with surface morphological features<br><br>- No clinical information |
| Hyun et al, 2016 [147] | NLP and CNN AlexNet and VGG16 | To achieve neonatal brain ultrasound scans in classifying and/or annotating neonatal using combination of NLP and CNN | 2372 de identified NS report | 11,205 NS head Images | 87%<br>(AUC) | + Automated labelling<br><br>- No clinical variable |
| Kim et al, 2022 [148] | CNN(VGG16) Transfer learning | To assesses whether a convolutional neural network (CNN) can be trained via transfer learning to accurately diagnose germinal matrix hemorrhage on head ultrasound | | 400 head ultrasounds<br>(200 with GMH,200 without hemorrhage) | 92% (AUC) | + First study to evaluate GMH with grade and saliency map<br>+ Not confirmed with MRI or labelling by radiologists<br>- Small sample size which limited the training, validation and testing of CNN algorithm |
| Li et al, 2021[149] | ResU-Net | Diffuse white matter abnormality (DWMA) on VPI's MR images at term-equivalent age | 98 VPI<br>28 VPI | 3 Tesla Brain MRI T1 and T2 weighted | 87.7%<br>(Dice Score)<br>92.3%<br>(accuracy) | + Developed to diffuse white matter abnormality on T2-weighted brain MR images of very preterm infants<br>+ 3D ResU-Net model achieved better DWMA segmentation performance than multiple peer deep learning models.<br>- Small sample size<br>- Limited clinical information |
| Greenbury et al, 2021 [150] | | To acquire understanding | n=45,679) over a six-year period | EHR | | + Identifying relationships |
| | Agnostic, unsupervised ML<br><br>Dirichlet Process Gaussian Mixture Model (DPGMM) | into nutritional practice, a crucial component of neonatal intensive care | UK National Neonatal Research Database (NNRD) | clustering on time analysis on daily nutritional intakes for extremely preterm infants born < 32 weeks gestation | | between nutritional practice and exploring associations between nutritional practices and outcomes using two outcomes: discharge weight and BPD<br>+Large national multi center dataset<br>- Strong likelihood of multiple interactions between nutritional components could be utilized in records |
| Ervural et al, 2021[151] | CNN<br>Data augmentation | To detect respiratory abnormalities of neonates by AI using limited thermal image | 34 neonates<br>680 images<br>2060 thermal images<br>(11 testing)<br>23 training) | Thermal camera image | 85%<br>(accuracy) | + CNN model and data enhancement methods were used to determine respiratory system anomalies in neonates.<br>- Small sample size<br>-There is no follow-up and no clinical information |
| Wang et al, 2018 [152] | DCNN | To classify automatically and grade a retinal hemorrhage | 3770<br>newborns with retinal hemorrhage of different severity (grade 1, 2 and 3) and normal controls from a large cross-sectional investigation in China. | 48,996 digital fundus images | 97.85% to 99.96%<br>(accuracy)<br><br>98.9% -100%<br>AUC) | +The first study to show that a DCNN can detect and grade neonatal retinal hemorrhage at high performance levels |
| Brown et al,2018 [153] | DCNN | To develop and test an algorithm based on DL to automatically diagnose plus disease from retinal photographs | 5511 retinal photographs (trained)<br>independent set of 100 images | Retinal images | 94%<br>(AUC)<br>98%<br>(AUC) | + Outperforming 6 of 8 ROP expert<br>+ Completely automated algorithm detected plus disease in ROP with the same or greater accuracy as human doctors<br>+ Disease detection, monitoring, and prognosis in ROP-prone neonates<br>- No clinical information and no clinical variables |
| Wang et al, 2018 [154] | DNN (ID-Net Gr-Net) | To automatically develop identification and grading system from retinal fundus images for ROP | 349 cases for identification<br><br>222 cases for grading | Retinal fundus images | Id-Net : 96.64% (sensitivity) 99.33% (specificity) 99.49% (AUC)<br><br>Gr-Net: 88.46% (sensitivity)<br><br>92.31% (specificity)<br><br>95.08% (AUC) | + Large dataset including training, testing and ,comparison with human experts.<br><br>+ Good example of human in the loop models<br><br>+ Code is available<br><br>- No clinical grading included<br><br>- Dataset is not available |
| Taylor et al,2019 [155] | DCNN Quantitative score | To describe a quantitative ROP severity score derived using a DL algorithm designed to evaluate plus disease and to assess its utility for objectively monitoring ROP progression | Retinal images | 871 premature infants | | + ROP vascular severity score is related to disease category at a specific period and clinical course of ROP in preterm<br>- Retrospective cohort study<br>- No follow-up for patients<br>- Low generalizability |
| Campbell et al,2021 [156] | DL(U-Net) Tensor Flow ROP Severity Score(1-9) | Evaluate the effectiveness of artificial intelligence (AI)–based screening in an Indian ROP telemedicine program and whether differences in ROP severity between neonatal care units (NCUs) identified by using AI are related to differences in oxygen-titrating capability | 4175 unique images from 1253 eye examinations retinopathy of Prematurity Eradication Save Our Sight ROP telemedicine program | 363 infants from 32 NCUs | 98% (AUC) | + Integration of AI into ROP screening programs may lead to improved access to care for secondary prevention of ROP and may facilitate assessment of disease epidemiology and NCU resources |
| Xu et al,2021 [157] | -Wireless sensors<br>-Pediatric focused algorithm | To enhance monitoring with wireless sensors | | By the middle of 2021, there were 15,000 pregnant women and up to 500 newborns. | | + Future predictive algorithms of clinical outcomes for neonates |
| | -ML and data analytics<br>-cloud based dashboards | | | 1000 neonates | | + As small as 4.4 cm 2.4 cm and as thin as 1 mm in totally wirelessly powered versions, these devices provide continuous monitoring in this sensitive group |
| Werth et al,2019 [158] | Sequential CNN ResNet | Automated sleep state requirement without EEG monitoring | 34 stable preterm infants | Vital signs were recorded<br>ECG R peaks were analyzed | Kappa of 0.43 ± 0.08<br>Kappa of 0.44 ± 0.01<br>Kappa of 0.33 ± 0.04 | +Non invasive sleep monitoring from ECG signals<br>-Retrospective study<br>- Video were not used in analysis |
| Ansari et al,2022 [159] | A Deep Shared Multi-Scale Inception Network | Automated sleep detection with limited EEG Channels | 26 preterm infants | 96 longitudinal EEG recordings | Kappa 0.77 ± 0.01 (with 8-channel EEG) and 0.75 ± 0.01 (with a single bipolar channel EEG | + The first study using Inception-based networks for EEG analysis that utilizes filter sharing to improve efficiency and trainability.<br>+ Even a single EEG channel making it more practical<br>- Small sample size<br>- Retrospective<br>- No clinical information |
| Ansari et al,2018 [160] | CNN | To discriminate quiet sleep from nonquiet sleep in preterm infants (without human labelling and annotation) | 26 preterm infants | 54 EEG recordings for training<br>43 EEG recording for the test<br>(at 9 and 24 months corrected age, a normal neurodevelopmental outcome score (Bayley Scales of Infant Development-II, mental and motor score >85)) | 92% (AUC)<br>98% (AUC) | + CNN is a viable and rapid method for classifying neonatal sleep phases in preterm babies<br>+ Clinical information<br>- Retrospective<br>- The paucity of EEG recordings below 30 weeks and beyond 38 weeks postmenstrual age<br>- Lack of interpretability of the features |
| Moeskops et al 2017[161] | CNN for MRI segmentation [162]<br><br>SVM for neurocognitive outcome prediction | To predict cognitive and motor outcome at 2-3 years of preterm infants from MRI at 30th and 40th weeks of PMA | 30 weeks (n=86)<br>40 weeks ( n=153) | 3 T Brain MRI at 30th and 40th weeks of PMA<br><br>BSID-III at average age of 29 months (26-35) | Cognitive Outcome (BSID<85)<br><br>78% (AUC) 30 weeks of PMA<br><br>70% (AUC) 40 weeks of PMA<br><br>Motor Outcome BSID< 85<br><br>80% (AUC) 30 weeks of PMA<br><br>71% (AUC) 40 weeks of PMA | + Brain MRI can predict cognitive and motor outcome<br><br>+ Segmentations, quantitative descriptors, classification were performed and<br><br>+ Volumes, measures of cortical morphology were included as a predictor<br><br>- Small sample size<br><br>- Retrospective design |

**Table 5:** DL based studies in neonatology using imaging and non-imaging for prediction.

| Study | Approach | Purpose | Dataset | #Non-Image Data | #-Image data | AUC/ accuracy | Pros(+)<br>Cons(-) |
|---|---|---|---|---|---|---|---|
| Saha et al, 2020 [163] | CNN | To predict abnormal motor outcome at 2 years from early brain diffusion magnetic resonance imaging (MRI) acquired between 29 and 35 weeks postmenstrual age (PMA) | 77 very preterm infants (born <31 weeks gestational age (GA)) | At 2 years CA, infants were assessed using the Neuro-Sensory Motor Developmental Assessment (NSMDA) | 3 T brain diffusion MRI | 72% (AUC) | + Neuromotor outcome can be predicted directly from very early brain diffusion MRI (scanned at ~30 weeks PMA), without the requirement of constructing brain connectivity networks, manual scoring, or predefined feature extraction<br>+ Cerebellum and occipital and frontal lobes were related motor outcome<br><br>-Small sample size |
| Shabanian et al, 2019 [164] | Based on MRIs, the 3D CNN algorithm can promptly and accurately diagnose | Neurodevelopmental age estimation | 112 individuals | | 1.5T MRI from NIMH Data Achieve | 95% (accuracy)<br>98.4% (accuracy) ( | + 3D CNNs can be used to accurately estimate neurodevelopmental age in infants based on brain MRIs |
| | neurodevelopmental age | | | | | | - Restricted clinical information<br>- No clinical variable<br>- Small sample size which limited the training, validation and testing of CNN algorithm |
| He et al, 2020 [165] | Supervised and unsupervised learning | In terms of predicting abnormal neurodevelopmental outcomes in extremely preterm newborns, multi-stage DTL(deep transfer learning) outperforms single-stage DTL. | 33 preterm infants<br><br>Retrained in 291 neonates | Bayley Scales of Infant and Toddler Development III at 2 years corrected age | 3 Tesla Brain MRI T1 and T2 weighted | 86% (cognitive deficit-AUC)<br><br>66% (language deficit-AUC)<br>84% (motor deficit-AUC) | + Risk stratification at term-equivalent age for early detection of long-term neurodevelopmental abnormalities and directed earlier therapies to enhance clinical outcomes in extremely preterm infants<br>- The investigation of the brain's functional connectome was based on an anatomical/structural atlas as opposed to a functional brain parcellated atlas. |
| Temple et al, 2016 [166] | supervised ML and NLP | To identify patients that will be medically ready for discharge in the subsequent 2–10 days. | 4,693 patients (103,206 patient-days | NLP using a bag of words (BOW) surgical diagnoses, pulmonary hypertension, retinopathy of prematurity, and psychosocial issues | | 63.3%( AUC)<br>67.7% (AUC)<br>75.2% (AUC)<br>83.7% (AUC) | + Could potentially avoid over 900 (0.9%) hospital days |

### ML Applications in Neonatal Mortality

Neonatal mortality is a major factor in child mortality. Neonatal fatalities account for 47 percent of all mortality in children under the age of five, according to the World Health Organization[167]. It is, therefore, a priority to minimize worldwide infant mortality by 2030[126,168].

ML investigated infant mortality, its reasons, and its mortality prediction[113,116,126,127,130,131,169]. In a recent review, 1.26 million infants born from 22 weeks to 40 weeks of gestational age were enrolled[169]. Predictions were made as early as 5 minutes of life and as late as 7 days. An average of four models per investigation were neural networks, random forests, and logistic regression (58.3 %)[169]. Two studies (18.2%) completed external validation, although five (45.5%) published calibration plots[169]. Eight studies reported AUC, and five supplied sensitivity and specificity[169]. The AUC was 58.3% - 97.0%[169]. Sensitivities averaged 63% to 80%, and specificities 78% to 98%[169]. Linear regression analysis was the best overall model despite having 17 features[169]. This analysis highlighted the most prevalent AI neonatal mortality measures and predictions. Despite the advancement in neonatal care, it is crucial that preterm infants remain highly susceptible to mortality due to immaturity of organ systems and increased susceptibility to early and late sepsis[170]. Addressing these permanent risks necessitates the utilization of ML to predict mortality[113,116,127,130,131,135]. Early studies employed ANN and fuzzy linguistic models and achieved an AUC of 85-95% and accuracy of 90%[113,126]. New studies in a large preterm populations and extremely low birthweight infants found an AUC of 68.9 - 93.3%[130,137]. There are some shortcomings in these studies; for example, none of them used vital parameters to represent dynamic changes, and hence, there was no improvement in clinical practice in neonatology. Unsurprisingly, gestational age, birthweight, and APGAR scores were shown as the most important variables in the models[116,136]. Future research is suggested to focus on external evaluation, calibration, and implementation of healthcare applications[169].

Neonatal sepsis, which includes both early onset sepsis and late onset sepsis, is a significant factor contributing to neonatal mortality and morbidity[171]. Neonatal sepsis diagnosis and antibiotic initiation present considerable obstacles in the field of neonatal care, underscoring the importance of implementing comprehensive interventions to alleviate their profound negative consequences. The studies have predicted early sepsis from heart rate variability with an accuracy of 64 - 94%[123]. Another secondary analysis of multicenter data revealed that clinical biomarkers weighed the ML decision by integrating all clinical and lab variables and achieved an AUC of 73-83%[138].

**ML Applications in Neurodevelopmental Outcome**

Recent advancements in neonatal healthcare have resulted in a decrease in the incidence of severe prenatal brain injury and an increase in the survival rates of preterm babies[172]. However, even though routine radiological imaging does not reveal any signs of brain damage, this population is nonetheless at significant risk of having a negative outcome in terms of neurodevelopment[173-176]. It is essential to discover early indicators of abnormalities in brain development that might serve as a guide for the treatment of preterm children at a greater risk of having negative neurodevelopmental consequences[177,178].

The most common reason for neurodevelopmental impairment is intraventricular hemorrhage (IVH) in preterm infants[179]. Two studies predicted IVH in preterm infants. Both studies have not deployed the ultrasound images in their analysis, they only predicted IVH according to the clinical variables[87,132].

Morphological studies have demonstrated that preterm birth is linked to smaller brain volume, cortical folding, axonal integrity, and microstructural connectivity[180,181]. Studies concentrating on functional markers of brain maturation, such as those derived from resting-state functional connectivity (rsFC) analyses of blood-oxygen-level dependent (BOLD) fluctuations, have revealed further impacts of prematurity on the developing connectome, ranging from decreased network-specific connectivity[63,178,182]. Many studies investigated brain connectivity in preterm infants[62,63,66,183] and brain structural analysis in neonates[64] and neonatal brain segmentation[67] with the help of ML methods. Similarly, one of the most important outcomes of neurodevelopment at 2-year-old-age is neurocognitive evaluations. The studies evaluated the morphological changes in the brain in relation to neurocognitive outcome[109-111] and brain age prediction[146,184]. It has been found that near-term regional white matter (WM) microstructure on diffusion tensor imaging (DTI) predicted neurodevelopment in preterm infants using exhaustive feature selection with cross-validation[110] and multivariate models of near-term structural MRI and WM microstructure on DTI might help identify preterm infants at risk for language impairment and guide early intervention[109,111] (**Table 4**). One of the studies that evaluated the effects of PPAR gene activity on brain development with ML methods[65] revealed a strong association between abnormal brain connectivity and implicating PPAR gene signaling in abnormal white matter development. Inhibited brain growth in individuals exposed to early extrauterine stress is controlled by genetic variables, and PPARG signaling has a formerly unknown role in cerebral development[65] **(Table 2)**.

Alternative to morphological studies, *neuromonitorization* is shown to be an important tool for which ML methods have been frequently employed, for example, in automatic seizure detection from video EEG[70,71,83,95] and EEG biosignals in infants and neonates with HIE[78,79,99,101,102]. The detection of artifacts[98,103], sleep states[95], rhythmic patterns[96], burst suppression in extremely preterm infants[97,100] from EEG records were studied with ML methods. EEG records are often used for HIE grading[82] too. It has been shown in those studies that EEG recordings of different neonate datasets found an AUC of 89% to 96%[78-80], accuracy 78%-87%[81,82] regarding seizure detection with different ML methods **(Table 3).**

**ML Applications in Predictions of Prematurity Complications (BPD, PDA and ROP)**

Another important cause of mortality and morbidity in the NICU is PDA (Patent Ductus Arteriosus). The ductus arteriosus is typically present during the fetal stage, when the circulation in the lungs and body is regularly supplied by the mother; in newborns, the ductus arteriosus closes functionally by 72 hours of age[185]. 20–50% of infants with a gestational age (GA) 32 weeks have the ductus arteriosus on day 3 of life[186], while up to 60% of neonates with a GA 29 weeks have the ductus arteriosus. The presence of PDA in preterm neonates is associated with higher mortality and morbidity, and physicians should evaluate if PDA closure might enhance the likelihood of survival vs. the burden of adverse effects[187-190].

ML methods were utilized on PDA detection from EHR[104] and auscultation records[105] such that 47 perinatal factors were analyzed with 5 different ML methods in 10390 very low birth weight infants' predicted PDA with an accuracy of 76%[104] and 250 auscultation records were analyzed with XGBoost and found to have an accuracy of 74%[105] **(Table 3).**

Bronchopulmonary dysplasia (BPD) is a leading cause of infant death and morbidity in preterm births. While various biomarkers have been linked to the development of respiratory distress syndrome (RDS), no clinically relevant prognostic tests are available for BPD at birth[124]. There are ML studies aiming to predict BPD from birth[125,135], gastric aspirate content[124] and genetic data[93] and it has been shown that BPD could be predicted with an accuracy of up to 86% in the best-case scenario[135] (**Table 5**), analysis of responsible genes with ML could predict BPD development with an AUC of 90%[93] (**Table 3**) and combination of gastric aspirate after birth and clinical information analysis with SVM predicted BPD development with a sensitivity of 88%[124] (**Table 5**).

In relation to published studies in BPD with ML based predictions, long term invasive ventilation is considered one of the most important risk factors for BPD, nosocomial infections, and increased hospital stay. There are ML based studies aiming to predict extubation failure[90,91,122] and optimum weaning time[92] using long term invasive ventilation information. It has been shown in those studies that predicted extubation failure with an accuracy of 83,2% to 87%[90,91,122] (**Tables 2 and 3**).

Retinopathy of prematurity (ROP) is another area of interest in the application of machine learning in neonatology[191]. ROP is a serious complication of prematurity that affects the blood vessels in the retina and is a leading cause of childhood blindness in high and middle-income countries, including the United States, among very low-birthweight (1500 g), very preterm (28–32 weeks), and extremely preterm infants (less than 28 weeks) [191]. Due to a shortage of ophthalmologists available to treat ROP patients, there has been increased interest in the use of telemedicine and artificial intelligence as solutions for diagnosing ROP[191]. Some ML methods, such as Gaussian mixture models, were employed to diagnose and classify ROP from retinal fundus images in studies [69,70,191], and it has been reported that the i-ROP[69] system classified pre-plus and plus disease with 95% accuracy. This was close to the performance of the three individual experts (96%, 94%, and 92%, respectively), and much higher than the mean performance of 31 nonexperts (81%)[69] (**Table 2**).

**Other ML Applications in Neonatal Diseases**

EHR and medical records were featured in ML algorithms for the diagnosis of congenital heart defects[72], HIE (Hypoxic Ischemic Encephalopathy) [86], IVH (Intraventricular Hemorrhage) [87,132], neonatal jaundice[68,88], prediction of NEC (Necrotizing Enterocolitis) [114], prediction of neurodevelopmental outcome in ELBW (extremely low birth weight) infants[112,119,130], prediction of neonatal surgical site infections[129] and prediction of rehospitalization[134] (**Table 5**).

Electronically captured physiologic data are evaluated as signal data, and they were analyzed with ML to detect artefact patterns[94],late onset sepsis, [133] and predict infant morbidity[120]. Electronically captured vital parameters (respiratory rate, heart rate) of 138 infants (≤34 weeks' gestation, birth weight ≤2000 gram) in the first 3 hours of life predicted an accuracy of overall morbidity and an AUC of 91%[120] (**Table 5**).

In addition to physiologic data, clinical data up to 12 hours after cardiac surgery in HLHS (hypoplastic left heart syndrome) and TGA (transposition of great arteries) infants were analyzed to predict PVL (periventricular leukomalacia) occurrence after surgery[118]. The F-score results for infants with HLHS and those without HLHS were 88% and 100%, respectively[118] (**Table 5**). Voice records were used to diagnose respiratory phases in infant cry[73], to classify neonatal diseases in infant cry[74], and to evaluate asphyxia from infant cry voice records[85]. Voice records of 35 infants were analyzed with ANN, and accuracy was found 85%[74]. Cry records of 14 infants in their 1$^{st}$ year of life were analyzed with SVM and GMM, and phases of respiration and crying rate were quantified with an accuracy of 86%[73] (**Table 3**).

SVM was the most commonly used method in the diagnosis of metabolic disorders of newborns, including MMA (methylmalonic acidemia) [75], PKU (phenylketonuria) [76,77], MCADD (medium-chain acyl CoA dehydrogenase deficiency) [76]. During the Bavarian newborn screening program, dried blood samples were analyzed with ML and increased the positive predictive value for PKU (71.9% versus 16,2) and for MCADD (88.4% versus 54.6%)[76] (**Table 3**).

### Neonatology with Deep Learning

The main uses of DL in clinical image analysis are categorized into three categories: classification, detection, and segmentation. Classification involves identifying a specific feature in an image, detection involves locating multiple features within an image; and segmentation involves dividing an image into multiple parts[7,9,140,147-149,192-194].

### Neuroradiological Evaluation with AI in Neonatology

Neonatal neuroimaging can establish early indicators of neurodevelopmental abnormality to provide early intervention during a time of maximal neuroplasticity and fast cognitive and motor development[110,175]. DL methods can assist in an earlier diagnosis than clinical signs would indicate.

The imaging of an infant's brain using MRI can be challenging due to lower tissue contrast, substantial tissue inhomogeneities, regionally heterogeneous image appearance, immense age-related intensity variations, and severe partial volume impact due to the smaller brain size. Since most of the existing tools were created for adult brain MRI data, infant-specific computational neuroanatomy tools are recently being developed. A typical pipeline for early prediction of neurodevelopmental disorders from infant structural MRI (sMRI) is made up of three basic phases. (1) Image preprocessing, tissue segmentation, regional labeling, and extraction of image-based characteristics (2) Surface reconstruction, surface correspondence, surface parcellation, and extraction of surface-based features (3) Feature preprocessing, feature extraction, AI model training, and prediction of unseen subjects[195]. The segmentation of a newborn brain is difficult due to the decreased SNR (signal to noise ratio) resulting from the shorter scanning duration enforced by predicted motion restrictions and the diminutive size of the neonatal brain. In addition, the cerebrospinal fluid (CSF)-gray matter border has an intensity profile comparable to that of the mostly unmyelinated white matter (WM), resulting in significant partial volume effects. In addition, the high variability resulting from the fast growth of the brain and the continuing myelination of WM imposes additional constraints on the creation of effective segmentation techniques. Several non-DL-based approaches for properly segmenting newborn brains have been presented over the years. These methods may be broadly classified as parametric[196-1981], classification[199], multi-atlas fusion[200,201], and deformable models[145,202]. The Dice Similarity Coefficient metric is used for image segmentation evaluation; the higher the dice, the higher the segmentation accuracy[10] (**Table 1**).

In the NeoBrainS12 2012 MICCAI Grand-Challenge (**https://neobrains12.isi.uu.nl**), T1W and T2W images were presented with manually segmented structures to assess strategies for segmenting neonatal tissue[196]. Most methods were found to be accurate, but classification-based approaches were particularly precise and sensitive. However, segmentation of myelinated vs. unmyelinated WM remains a difficulty since the majority of approaches[196] failed to consistently obtain reliable results.

Future research in neonatal brain segmentation will involve a more thorough neural segmentation network. Current studies are intended to highlight efficient networks capable of producing accurate and dependable segmentations while comparing them to existing conventional computer vision techniques. In the perspective of comparing previous efforts on newborn brain segmentation, the small sample size of high-quality labeled data must also be recognized as a significant restriction[145]. The field of artificial intelligence in neonatology has progressed slowly due to a shortage of open-source algorithms and the availability of datasets.

Future research should also focus on improving the accuracy of DL for diagnosing germinal matrix hemorrhage and figuring out how DL can help a radiologist's workflow by comparing how well sonographers identify studies that look suspicious. More studies could also look at how well DL works for accurately grading germinal matrix hemorrhages and maybe even small hemorrhages that a radiologist can see on an MRI but not on a head ultrasound. This could be useful in improving the diagnostic capabilities of head ultrasound in various clinical scenarios[148].

**Evaluation of Prematurity Complications with DL in Neonatology**

In the above discussion, we have addressed the primary applications of DL in relation to disease prediction. These include DL for analyzing conditions such as PDA (patent ductus arteriosus) [140], IVH (intraventricular ventricular hemorrhage) [147,148], BPD (bronchopulmonary dysplasia) [150], ROP (retinopathy of prematurity) [153,155,156], retinal hemorrhage[152] diagnosis. This also includes DL applications for analyzing MR images [149,164] and combined with EHR data[163,165] for predicting neurocognitive outcome and mortality. Additionally, DL has potential applications in treatment planning and discharge from the NICU[203], including customized medicine and follow-up[6,124,169] (**Tables 6 and 7**).

Digital imaging and analysis with AI are promising and cost-effective tools for detecting infants with severe ROP who may need therapy[153-155,191]. Despite limitations such as image quality, interpretation variability, equipment costs, and compatibility issues with EHR systems, AI has been shown to be effective in detecting ROP[204]. Studies comparing BIO (Binocular Indirect Ophthalmoscope) to telemedicine have shown that both methods have equivalent sensitivity for identifying zone disease, plus disease, and ROP. However, BIO was found to be slightly better at identifying zone III and stage 3 ROP[205,206]. DL algorithms were applied to 5511 retinal images, achieving an AUC of 94% (diagnosis of normal) and 98% (diagnosis of plus disease), outperforming 6 out of 8 ROP experts[153]. In another study, DL was used to quantify the clinical progression of ROP by assigning ROP vascular severity scores[155]. A consecutive study with a large dataset showed in 4175 retinal images from 32 NICUs, resulting in an AUC of 98% for detecting therapy required ROP with DL[156]. The use of AI in ROP screening programs may increase access to care for secondary prevention of ROP and enable the evaluation of disease epidemiology[156] (**Table 6**).
Signal detection for sleep protection in the NICU is another ongoing discussion. DL has been used to analyze infant EEGs and identify sleep states. Interruptions of sleep states have been linked to problems in neuronal development[207]. Automated sleep state detection from EEG records[159,160] and from ECG monitoring parameters[158] were demonstrated with DL. The underperformance of the all-state classification (kappa score 0.33 to 0.44) was likely owing to the difficulties in differentiating small changes between states and a lack of enough training data for minority classes[158] (**Table 6**). DL has been found to be effective in real-time evaluation of cardiac MRI for congenital heart disease[139]. Studies have shown that DL can accurately calculate ventricular volumes from images rebuilt using residual UNet, which are not statistically different from the gold standard, cardiac MRI. This technology has the potential to be particularly beneficial for infants and critically ill individuals who are unable to hold their breath during the imaging process[139] (**Table 6**).

DL-based 3D CNN algorithms have been used to demonstrate the automated classification of brain dysmaturation from neonatal brain MRI[144]. In a study, brain MRIs of 90 term neonates with congenital heart diseases and 40 term healthy controls were analyzed using this method, which achieved an accuracy of 98%. This technique could be useful in detecting brain dysmaturation in neonates with congenital heart diseases[144] (**Table 6**).

DL algorithms have been used to classify neonatal diseases from thermal images[141-143,151]. These studies analyzed neonatal thermograms to determine the health status of infants and achieved good AUC scores[141-143,151]. However, these studies didn't include any clinical information (**Table 6**).

Two large scale studies showed breakthrough results regarding the effect of nutrition practices in NICU[150] and wireless sensors in NICU[157]. A nutrition study revealed that nutrition practices were associated with discharge weight and BPD[150]. This exemplifies how unbiased ML techniques may be used to effectively bring about clinical practice changes[150]. Novel, wireless sensors can improve monitoring, prevent iatrogenic injuries, and encourage family-centered care[157]. Early validation results show performance equal to standard-of-care monitoring systems in high-income nations. Furthermore, the use of reusable sensors and compatibility with low-cost mobile phones may reduce monitoring.

**Discussions**

The studies in neonatology with AI were categorized according to the following criteria.

i) The studies were performed with ML or DL,

ii) imaging data or non-imaging data were used,

iii) according to the aim of the study: diagnosis or other predictions.

Most of the studies in neonatology were performed with ML methods in the pre-DL era. We have listed 12 studies with ML and imaging data for diagnosis. There are 33 studies that used non-imaging data for diagnosis purposes. Imaging data studies cover BA diagnosis from stool color[60], postoperative enteral nutrition of neonatal high intestinal obstruction[61], functional brain connectivity in preterm infants[62,65-67,178], ROP diagnosis[69,70], neonatal seizure detection from video records[71], newborn jaundice screening[68]. Non-imaging studies for diagnosis include the diagnosis of congenital heart defects[72], baby cry analysis[73,74,85], inborn metabolic disorder diagnosis and screening[75-77], HIE grading[79,82,86,99,108], EEG analysis[79,80,84,95-97,99-101,103,107,160], PDA diagnosis[104,105], vital sign analysis and artifact detection[94], extubation and weaning analysis[90-92,94], BPD diagnosis[93]. ML studies with imaging data for prediction are focused on neurodevelopmental outcome prognosis from brain MRIs [93,109-111,161,198]. ML based non-imaging data for prediction encompassed mortality risk[113,116,127,130], NEC prognosis[114], morbidity[120,131], BPD[124,125].

When it comes to DL applications, there has been less research conducted compared to ML applications. The focus of DL with imaging and non-imaging data focused on brain segmentation[144,145,149,164,165], IVH diagnosis[148], EEG analysis[159,160], neurocognitive outcome, [163] PDA and ROP diagnosis[153,155,156]. Upcoming articles and research will surely be from the DL field, though.

It is worth noting that there have also been several articles and studies published on the topic of the application of AI in neonatology. However, the majority of these studies do not contain enough details, are difficult to evaluate side-by-side, and do not give the clinician a thorough picture of the applications of AI in the general healthcare system [64,93,106,109-111,115,117-119,121,124,125,128,129,131,144-146,152,159,165,169,208].

There are several limitations in the application of AI in neonatology, including a lack of prospective design, a lack of clinical integration, a small sample size, and single center evaluations. DL has shown promise in bioscience and biosignals, extracting information from clinical images, and combining unstructured and structured data in EHR. However, there are some issues that limit the success of DL in medicine, which can be grouped into six categories. In the following paragraphs, we'll examine the key concerns related to DL, which have been divided into six components:

1) Difficulties in clinical integration, including the selection and validation of models;
2) the need for expertise in decision mechanisms, including the requirement for human involvement in the process;
3) lack of data and annotations, including the quality and nature of medical data; distribution of data in the input database; and lack of open-source algorithms and reproducibility;
4) lack of explanations and reasoning, including the lack of explainable AI to address the "black-box" problem;
5) lack of collaboration efforts across multi-institutions; and
6) ethical concerns [5,6,9,10,209].

**Difficulties in clinical integration**

Despite the accuracy that AI has reached in healthcare in recent years, there are several restrictions that make it difficult to translate into treatment pathways. First, physicians' suspicion of AI-based systems stems from the lack of qualified randomized clinical trials, particularly in the field of pediatrics, showing the reliability and/or improved effectiveness of AI systems compared to traditional systems in diagnosing neonatal diseases and suggesting appropriate therapies. The studies' pros and cons are discussed in tables and relevant sections. Studies are mainly focused on imaging based or signal based studies in terms of one variable or disease. Neonatologists and pediatricians need evidence-based proven algorithm studies. There are only six prospective clinical trials in neonatology with AI[107,210-212]. The one is detecting neonatal seizures with conventional EEG in the NICU which is supported by the European Union Cost Program in 8 European NICU[107]. Neonates with a corrected gestational age between 36 and 44 weeks who had seizures or were at high risk of having seizures and needed EEG monitoring were given conventional EEG with ANSeR (Algorithm for Neonatal Seizure Recognition) coupled with an EEG monitor that displayed a seizure probability trend in real time (algorithm group) or continuous EEG monitoring alone (non-algorithm group) [107]. The algorithm is not available, and the code is not shared. Another one is a study showing the physiologic effects of music in premature infants [211]. Even so, it could not be founded on any AI analysis in this study. The third study, "Rebooting Infant Pain Assessment: Using Machine Learning to Exponentially Improve Neonatal Intensive Care Unit Practice (BabyAI)," is newly posted and recruiting [212]. The fourth study, " Using sensor-fusion and machine learning algorithms to assess acute pain in non-verbal infants: a study protocol," aims to collect data from 15 subjects: preterm infants, term infants within the first month of age in NICU admission and their follow-up data at 3rd and 6th months of age. They record pain signals using facial electromyography(EMG), ECG, electrodermal activity, oxygen saturation, and EEG in real time, and they will analyze the data with ML methods to evaluate pain in neonates. The data is in iPAS (NCT03330496) and is updated as recruitment completed[213]. However, no result has been submitted. The fifth study, “Prediction of Extubation Readiness in Extreme Preterm Infants by the Automated Analysis of Cardiorespiratory Behavior: APEX study” [214] records revealed that the recruitment was completed in 266 infants. Still, no results have been released yet (NCT01909947). To sum up, there is only one prospective multicenter randomized AI study that has been published with its results.

There is an unmet need to plan clinically integrated prospective and real time data collection studies in neonatology. The clinical situation of infants changed rapidly, and real time designed studies would be significant by analyzing multimodal data and including imaging and non-imaging components.

**The need for expertise in the decision mechanisms**

In terms of neonatologists determining whether to implement a system's recommendation, it may be required for that system to present supporting evidence [109,110,124,208]. Many suggested AI solutions in the medical field are not expected to be an alternative to the doctor's decision or expertise but rather to serve as helpful assistance. When it comes to struggling neonatal survival without sequela, AI may be a game changer in neonatology. The broad range of neonatal diseases and different clinical presentations of neonates according to gestational age and postnatal age make accurate diagnosis even harder for neonatologists. AI would be effective for early disease detection and would assist clinicians in responding promptly and fostering therapy outcomes.

Neonatology has multidisciplinary collaborations in the management of patients, and AI has the potential to achieve levels of efficacy that were previously unimaginable in neonatology if more resources and support from physicians were allocated to it. Neonatology collaborates and closely works with other specialties of pediatrics, including perinatology, pediatric surgery, radiology, pediatric cardiology, pediatric neurology, pediatric infectious disease, neurosurgery, cardiovascular surgery, and other subspecialties of pediatrics. Those multidisciplinary workflows require patient follow-up and family involvement. AI based predictive analysis tools might address potential risks and neurologic problems in the future. AI supported monitoring systems could analyze real time data from monitors and detect changes simultaneously. These tools could be helpful not only for routine NICU care but also for "family centered care" [215,216] implications. Although neonatologists could be at the center of decision making and giving information to parents, AI could be actively used in NICUs. Hybrid intelligence would provide a follow-up platform for abrupt and subtle clinical changes in infants' clinical situations.

Given that many medical professionals have a limited understanding of DL, it may be difficult to establish contact and communication between data scientists and medical specialists. Many medical professionals, including pediatricians and neonatologists in our instance, are unfamiliar with AI and its applications due to a lack of exposure to the field as an end user. However, the authors also acknowledge the increasing efforts in building bridges among many scientists and institutions, with conferences, workshops, and courses, that clinicians have successfully started to lead AI efforts, even with software coding schools by clinicians[217-221].

Neonatal critical conditions will be monitored by the human in the loop systems in the near future, and AI empowered risk classification systems may help clinicians prioritize critical care and allocate supplies precisely. Hence, AI could not replace neonatologists, but there would be a clinical decision support system in the critical and calls for prompt response environment of NICU.

**Lack of imaging data and annotations and reproducibility problems**

There is a rising interest in building deep learning approaches to predict neurological abnormalities using connectome data; however, their usage in preterm populations has been limited[62,63,66,177,182]. Similar to most DL applications, the training of such models often requires the use of big datasets[11]; however, large neuroimaging datasets are either not accessible or difficult and expensive to acquire, especially in the pediatric world. Since the success of DL methods currently relies on well labeled data and high-capacity models requiring several iterative updates across many labeled examples and obtaining millions of labeled examples, is an extreme challenge, there is not enough jump in the neonatal AI applications.

As a side note, accurate labeling always requires physician effort and time, which overcomplicates the current challenges. Unfortunately, there is no established collaboration between physicians and data scientists at a large scale that can ease some of the challenges (data gathering/sharing and labeling). Nonetheless, once these problems are addressed, DL can be used in prevention and diagnosis programs for optimal results, radically transforming clinical practice. In the following, we envision the potential of DL to transform other imaging modalities in the context of neonatology and child health.

The requirement for a massive volume of data is a significant barrier, as mentioned earlier. The quantity of data needed by an AI or ML system can grow in proportion to the sophistication of its underlying architecture; deep neural networks(DNN), for example, have particularly high volume of data needs. It's not enough that the needed data just be sufficient; they also need to be of good quality in terms of data cleaning and data variability (both ANN and DNN tend to avoid overfitting data if the variability is high). It may be difficult to collect a substantial amount of clean, verified, and varied data for several uses in neonatology. For this reason, there is a data repository shared with neonatal researchers, including EHR [208] and clinical variables. Some approaches for addressing the lack of labeled, annotated, verified, and clean datasets include: (1) building and training a model with a very shallow network (only a few thousand parameters) and (2) data augmentation. Data augmentation techniques are not helpful in the medical imaging field or medical setting[222].

In the field of neonatal imaging, high-quality labeling and medical imaging data are exceedingly uncommon. One of the other comparable available neonatal data sets the authors are aware of has just ten individuals[200,223,224]. This pattern holds even in more recent research, as detailed by the majority of studies involving little more than 20 individuals [199]. Regardless of sample size and technology, it is crucial to be able to generalize to new data in the field of image segmentation, especially considering the wide range of MRI contrasts and variations between scanners and sequences between institutions. Moreover, it is generally known that models based on DL have weak generalization skills on unseen data. This is especially crucial for the future translation of research into reality since (1) there is a shift between images obtained in various situations, and (2) the model must be retrained as these images become accessible. Adopting a strategy of continuous learning is the most practical way to handle this challenge. This method involves progressively retraining deep models while preventing any virtual memory loss on previously viewed data sets that may not be available during retraining. This field of endeavor will advance[145].

Most of the studies did not release their algorithms as open source to the libraries. Even though algorithms are available, it should be known whether separate training and testing datasets exist. There is a strong expectation that studies should have clarified which validation method has been chosen. In terms of comparing algorithm success, reproducibility is a crucial point. Methodological bias is another issue with this system. Research is frequently based on databases and guidelines from other nations that may or may not have patient populations similar to ours[110]. A database that only contains data that is applicable to the specific problem that must be solved; however, obtaining the relevant information may be difficult due to the number of databases.

### Lack of explanations and reasoning

The *trustworthiness* of algorithms is another obstacle[225]. The most widely used deep learning models use a black-box methodology, in which the model simply receives input and outputs a prediction without explaining its thought process. In high-stakes medical settings, this can be dangerous. Some models, on the other hand, incorporate human judgment (human-in-the-loop) or provide *interpretability maps* or *explainability* layers to illuminate the decision-making process. Especially in the field of neonatology, where AI is expected to have a significant impact, this trustworthiness is essential for its widespread adoption.

### Lack of collaboration efforts (multi-institutions) and privacy concerns

New collaborations have been forged because of this information; early detection and treatment of diseases that affect children, who make up a large portion of the world's population, will change treatment and follow-up status. Monitoring systems and knowing mortality and treatment activity with multi-site data will help. Considering the necessity for consent to the processing of personal health data by AI systems as an example of a subject related to the protection of privacy and security[110]. Efforts involving multiple institutions can facilitate training, but there are privacy concerns associated with the cross-site sharing of imaging data. Federated learning (FL) was introduced recently to address privacy concerns by facilitating distributed training without the transfer of imaging data [226]. Existing FL techniques utilize conditional reconstruction models to map from under sampled to fully-sampled acquisitions using explicit knowledge of the accelerated imaging operator[226]. Nevertheless, the data from various institutions is typically heterogeneous, which may diminish the efficacy of models trained using federated learning. *SplitAVG* is proposed as a novel heterogeneity-aware FL method to surmount the performance declines in federated learning caused by data heterogeneity[227].

**AI Ethics**

While AI has great promise for enhancing healthcare, it also presents significant ethical concerns. Ethical concerns in health AI include informed consent, bias, safety, transparency, patient privacy, and allocation, and their solutions are complicated to negotiate[228]. In neonatology, crucial decision-making is frequently accompanied by a complicated and challenging ethical component. Interdisciplinary approaches are required for progress[229]. The border of viability, life sustaining treatments[230] and the different regulations worldwide made AI utilization in neonatology more complicated. How an ethics framework is implemented in an AI in neonatology has not been reported yet, and there is a need for transparency for trustworthy AI.

The applications of AI in real-world contexts have the potential to result in a few potential benefits, including increased speed of execution; potential reduction in costs, both direct and indirect; improved diagnostic accuracy; increased healthcare delivery efficiency ("algorithms work without a break"); and the potential of supplying access to clinical information even to persons who would not normally be able to utilize healthcare due to geographic or economic constraints[4].

To achieve an accurate diagnosis, it is planned to limit the number of extra invasive procedures. New DL technologies and easy-to-implement platforms will enable regular and complete follow-up of health data for patients unable to access their records owing to a physician shortage, hence reducing health costs.

The future of neonatal intensive care units and healthcare will likely be profoundly impacted by AI. This article's objective is to provide neonatologists in the AI era with a reference guide to the information they might require. We defined AI, its levels, its techniques, and the distinctions between the approaches used in the medical field, and we examined the possible advantages, pitfalls, and challenges of AI. While also attempting to present a picture of its potential future implementation in standard neonatal practice. AI and pediatrics require clinicians' support, and due to the fact that AI researchers with clinicians need to work together and cooperatively. As a result, AI in neonatal care is highly demanded, and there is a fundamental need for a human (pediatrician) to be involved in the AI-backed up applications, in contrast to systems that are more technically advanced and involve fewer healthcare professionals.

## Methods

### Literature review and search strategy

We used PubMed™, IEEEXplore™, Google Scholar™, and ScienceDirect™ to search for publications relating to AI, ML, and DL applications towards neonatology. We have done a varying combination of the keywords( i.e., one from technical keywords and one from clinical keywords) for the search. Clinical keywords were "infant," "neonate," "prematurity," "preterm infant," "hypoxic ischemic encephalopathy," "neonatology," "intraventricular hemorrhage," "infant brain segmentation," "NICU mortality," "infant morbidity," " bronchopulmonary dysplasia," "retinopathy of prematurity." The inclusion criteria were (i) publication date between 1996-2022 and, (ii) being an artificial intelligence in neonatology study, (iii) written in English, (iv) published in a scholarly peer-reviewed
journal, and (v) conducted an assessment of AI applications in neonatology objectively. Technical keywords were AI, DL, ML, and CNN. Review papers, commentaries, letters to the editor and papers with only technical improvement without any clinical background, animal studies, and papers that used statistical models like linear regression, studies written in any language other than English, dissertation thesis, posters, biomarker prediction studies, simulation-based studies, studies with infants are older than 28 days of life, perinatal death, and obstetric care studies were excluded. The preliminary investigation yielded a substantial collection of articles, amounting to approximately 9000 in total. Through a meticulous examination of the abstracts of the papers, a subset of 987 research was found (Figure 2). Ultimately, 106 studies were selected for inclusion in our systematic review (Supplementary file). The evaluation encompassed diverse aspects, including sample size, methodology, data type, evaluation metrics, advantages, and limitations of the studies (Tables 2-7).

## Data Availability

Dr. E. Keles and Dr. U. Bagci have full access to all the data in the study and take responsibility for the integrity of the data and the accuracy of the data analysis. All study materials are available from the corresponding author upon reasonable request.

## Acknowledgement

This work is partially supported by the NIH NCI funding: R01-CA246704 and R01-CA240639.
Dr. E Keles is working as a senior clinical research associate in the Machine and Hybrid Intelligence Lab at the Northwestern University Feinberg School of Medicine, Department of Radiology.

Dr. U Bagci is director of the Machine and Hybrid Intelligence Lab and Associate Professor at the Department of Radiology, Northwestern University, Feinberg School of Medicine.

## Ethics Declaration

Dr. E. Keles has no COI.
Dr. U. Bagci discloses Ther-AI LLC.

## Author Contributions

Both authors contributed to the review design, data collection, interpretation of the data, analysis of data and drafting the report.

## Supplementary information

### Full Search Strategy and Bias Analysis

We used PubMed™, IEEEXplore™, Google Scholar™, and ScienceDirect™ to search for publications relating to AI, ML, and DL applications towards neonatology. We have done a varying combination of the keywords( i.e., one from technical keywords and one from clinical keywords) for the search. Clinical keywords were “infant,” “neonate,” “prematurity,” “preterm infant,” “hypoxic ischemic encephalopathy," "neonatology," "intraventricular hemorrhage," "infant brain segmentation," "NICU mortality," "infant morbidity," " bronchopulmonary dysplasia," "retinopathy of prematurity." The inclusion criteria were (i) publication date between 1996-2022 and, (ii) being an artificial intelligence in neonatology study, (iii) written in English, (iv) published in a scholarly peer-reviewed journal, and (v) conducted an assessment of AI applications in neonatology objectively. Technical keywords were AI, DL, ML, and CNN. Review papers, commentaries, letters to the editor and papers with only technical improvement without any clinical background, animal studies, and papers that used statistical models like linear regression, studies written in any language other than English, dissertation thesis, posters, biomarker prediction studies, simulation-based studies, studies with infants are older than 28 days of life, perinatal death, and obstetric care studies were excluded. An electronic reference manager (EndNote version 20) was utilized for reference organization. The article selection process involved two authors who independently performed the selection in two distinct phases, preceded by a pilot training test. In the initial phase, an assessment of titles and abstracts was carried out, alongside the application of predefined eligibility criteria. Subsequently, during the second phase, a thorough examination of full-text articles was undertaken by the reviewers, consistently aligning with the predetermined eligibility standards. Instances of variance were resolved through mutual agreement between the two authors. Following the first literature searches, each study's title and abstract were examined, and subsequently, studies that appeared to be possibly relevant were further evaluated for eligibility. The PRISMA flow diagram (Figure 2) contains comprehensive details regarding the study selection procedure. The preliminary investigation yielded a substantial collection of articles, amounting to approximately 9000 in total. To ensure accuracy and pertinence, we implemented a systematic and methodical procedure to carefully evaluate and choose publications that closely corresponded to our research objectives, study methodology, and the topic under investigation by following PRISMA 2020 guidelines[56]. Through a meticulous examination of the abstracts of the papers, a subset of 987 research was found (Figure 2). Ultimately, 106 studies were selected for inclusion in our systematic review. The evaluation encompassed diverse aspects, including sample size, methodology, data type, evaluation metrics, advantages, and limitations of the studies (Tables 2-7).

The included articles were assessed by both authors independently using the revised Cochrane risk-of-bias tool for non-randomized studies and were categorized into low risk, some concerns, or high risk. The risk of bias in the included studies was further evaluated using the QUADAS-2 (Quality Assessment of Diagnostic Accuracy Studies 2) tool[57-59]. The formal investigation of heterogeneity using meta-analysis was not possible due to the limited data availability. Additionally, the review protocol was not registered due to the same restriction.